\documentclass[journal]{IEEEtran}
\usepackage{amsmath}
\usepackage[utf8]{inputenc}
\usepackage{graphicx}

\usepackage{graphicx}
\usepackage{multirow}
\usepackage{booktabs}
\usepackage{amssymb}
\usepackage[colorlinks,linkcolor=blue]{hyperref}
\usepackage[table]{xcolor}  
\usepackage[numbers,sort&compress]{natbib}
\usepackage{threeparttable}

\ifCLASSINFOpdf

\else

\fi

\begin{document}

\title{Tortho-Gaussian: \\Splatting True Digital Orthophoto Maps}

\author{ Xin Wang,~\IEEEmembership{Member,~IEEE}, Wendi Zhang, Hong Xie, Haibin Ai, Qiangqiang Yuan,~\IEEEmembership{Member,~IEEE} and Zongqian Zhan,~\IEEEmembership{Member,~IEEE}
\thanks{Xin Wang and Wendi Zhang are co-first authors with equal contribution and importance. This work was supported by the National Natural Science Foundation of China (No. 42301507) and Natural Science Foundation of Hubei Province, China (No. 2022CFB727). (\textit{Corresponding author: Zongqian Zhan}) }
\thanks{Xin Wang, Wendi Zhang, Hong Xie, Qiangqiang Yuan and Zongqian Zhan are with the School of Geodesy and Geomatics, Wuhan University, Wuhan, 430079, China PR. (xwang@sgg.whu.edu.cn, 2023202140056@whu.edu.cn, hxie@sgg.whu.edu.cn, yqiang86@gmail.com,  zqzhan@sgg.whu.edu.cn).

Haibin Ai is with the Chinese Academy of Surveying $\&$ Mapping, Beijing, 100830, China PR. 
 (aihb@casm.ac.cn)}
\thanks{Manuscript received April 19, 2005; revised A\textit{}ugust 26, 2015.}}

\markboth{Journal of \LaTeX\ Class Files,~Vol.~14, No.~8, August~2015}%
{Shell \MakeLowercase{\textit{et al.}}: Bare Demo of IEEEtran.cls for IEEE Journals}

\maketitle

\begin{abstract}
True Digital Orthophoto Maps (TDOMs) are essential products for digital twins and Geographic Information Systems (GIS). Traditionally, TDOM generation involves a complex set of traditional photogrammetric process, which may deteriorate due to various challenges, including inaccurate Digital Surface Model (DSM), degenerated occlusion detections, and visual artifacts in weak texture regions and reflective surfaces, etc. To address these challenges, we introduce TOrtho-Gaussian, a novel method inspired by 3D Gaussian Splatting (3DGS) that generates TDOMs through orthogonal splatting of optimized anisotropic Gaussian kernel. More specifically, we first simplify the orthophoto generation by orthographically splatting the Gaussian kernels onto 2D image planes, formulating a geometrically elegant solution that avoids the need for explicit DSM and occlusion detection. Second, to produce TDOM of large-scale area, a divide-and-conquer strategy is adopted to optimize memory usage and time efficiency of training and rendering for 3DGS. Lastly, we design a fully anisotropic Gaussian kernel that adapts to the varying characteristics of different regions, particularly improving the rendering quality of reflective surfaces and slender structures. Extensive experimental evaluations demonstrate that our method outperforms existing commercial software in several aspects, including the accuracy of building boundaries, the visual quality of low-texture regions and building facades. These results underscore the potential of our approach for large-scale urban scene reconstruction, offering a robust alternative for enhancing TDOM quality and scalability. Project Web: \href{https://gwen233666.github.io/Ortho-Gaussian/}{https://gwen233666.github.io/Ortho-Gaussian/}
\end{abstract}


\IEEEpeerreviewmaketitle

\section{Introduction}
\begin{figure}[htbp]
\centering
\includegraphics[width=0.4\textwidth]{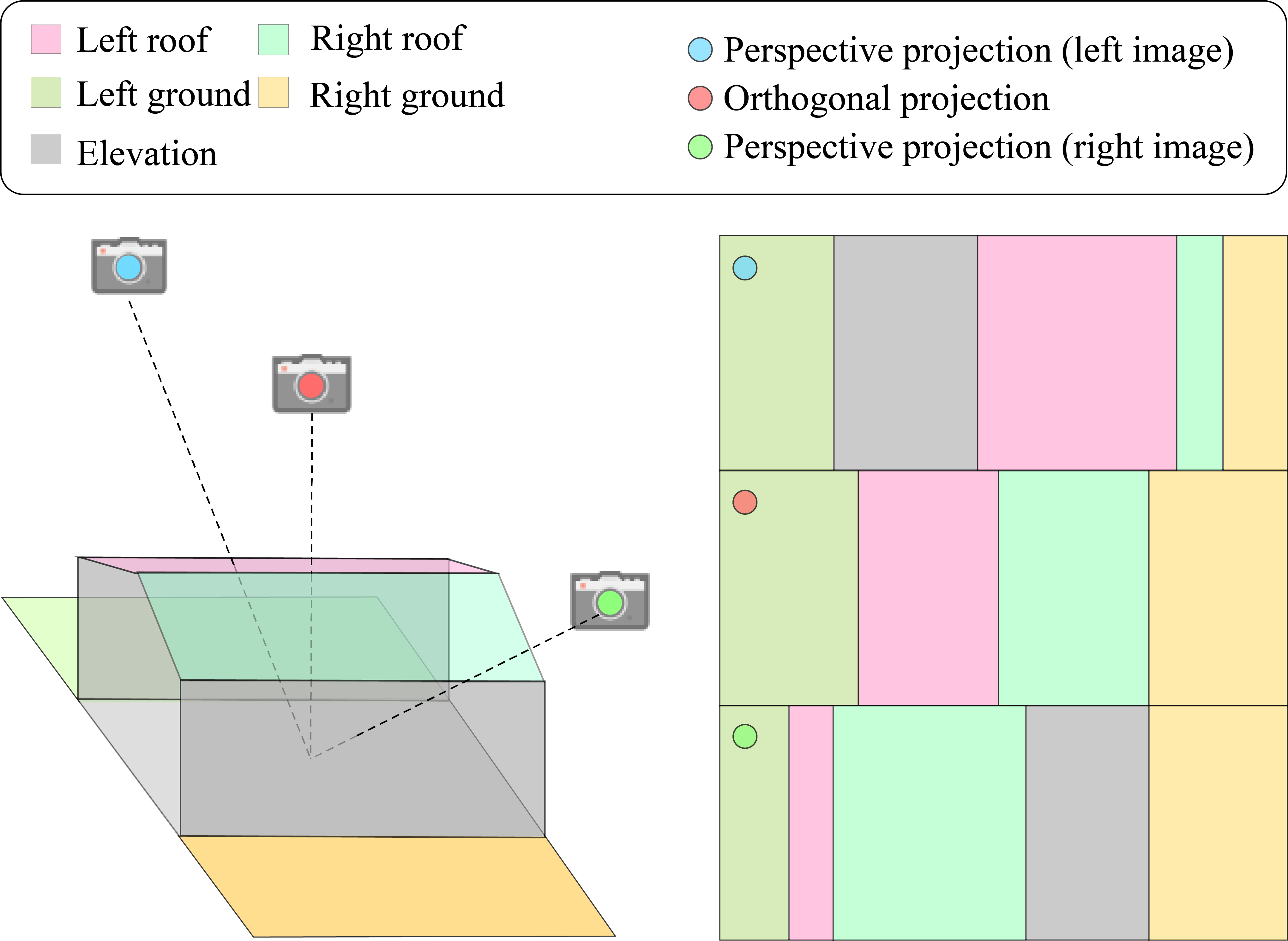}
\caption{A simplified model of building. It includes roofs and ground marked with different colors. The blue and green cameras represent perspective projection, with their imaging containing the gray building facades. The red camera represents orthographic projection.}
\label{figfecade}
\end{figure}

\IEEEPARstart{D}{igital} Orthophoto Maps (DOM) not only encompass rich texture information but also exhibit the geometric properties inherent to maps, making them highly applicable in various fields\cite{PE&RS1}, such as urban planning and cultural heritage preservation. Traditional DOMs generation takes oriented images and Digital Elevation Models (DEM) as input and employs digital differential rectification techniques. The generated result is a orthogonal nadir view of the surface that eliminates pin-hole projection distortions caused by terrain fluctuations and oblique photography, ensuring that the corresponding measuring scale is uniform throughout \cite{01dewitt2000elements}. DOM exhibits artifacts and incorrect geometries that are resulted from occlusion by the façades of man-made architectures, as Fig. \ref{figfecade} shows. Therefore, the TDOM (True Digital Orthophoto Maps) is more extensively used, which takes DSM into account and perform visibility check of mesh triangles for detecting occlusion \cite{02liu2018generating,03li2020review,PE&RS2}, and the obtained maps is less stemmed from the facades of buildings. 

Over the past few decades, numerous traditional methods for True Digital Orthophoto Map (TDOM) generation have been developed to address occlusion detection. One of the most widely used techniques is Z-buffering \cite{04amhar1998generation,05rau2000hidden,07uchidatriangle,08zhou2005comprehensive,09habib2006true,10bang2007comparative,11chen2007occlusion,12habib2007new,13antequera2008,14bang2010new}, which records the distances between the perspective center and object points corresponding to image pixels, thereby determining visibility by selecting the nearest points. Habib et al. \cite{12habib2007new} introduced the Angle-Based method, employing adaptive radial and spiral sweeps to analyze the angles between lines connecting the perspective center and Digital Surface Model (DSM) cells. They also proposed a Height-Based method, which assesses ray height against ground points along the search path, identifying occlusions when any ground point exceeds the ray height. Kuzmin et al. \cite{15kuzmin2004polygon} presented a Polygon-Based method that projects building polygons from the Digital Building Model (DBM) onto images, thereby enhancing orthographic image selection in sheltered areas. Wang et al. \cite{16wang2004method} applied a global variational model and texture matching techniques to fill in incomplete image data, thereby improving visual interpretation. Zhou et al. \cite{17zhou2016occlusion} developed a model linking ghost images to occlusions and employed seed growth algorithms for ghost detection.

In last few years, to the best of our knowledge, only a limited learning-based methods in scope have been proposed to enhance TDOM generation. Ebrahimikia and Hosseininaveh \cite{30zitnick2000cooperative} employ a learning-based approach to detect building edges in images and estimate corresponding 3D edge points to modify the DSM. They subsequently propose urban-SnowflakeNet for completing point cloud of edge regions \cite{31zhang2012true}. In addition, some works \cite{18shin2020true,19shin2021true} leverage the unoccluded properties of LiDAR intensity data within projection geometry to directly train a GAN (Generative Adversarial Network) model for generating TDOM. However, the first two methods suffer from limited generalization capabilities, while the latter heavily relies on intensity of LiDAR data. Furthermore, the emergence of neural radiance fields (NeRF) has revolutionized differentiable rendering by introducing an implicit 3D scene representation through Multi-layer Perceptrons (MLPs) \cite{34gilani2016automatic,35haggag2018towards,36wang2018true,37li2019fusion,38chen2018optimal,39yang2021efficient,40zhou2020urban,41yuan2023fully}, presenting an alternative approach for TDOM generation. Chen et al. \cite{62liu2025citygaussian} introduce the Ortho-Nerf, which utilizes Plenoxels \cite{20fridovich2022plenoxels} (an accelerated version of NeRF), and a true-ortho-volume rendering strategy is employed for generating TDOM. Qu et al. \cite{62liu2025citygaussian} take satellite images as input and apply the RPC (Rational polynomial coefficients) instead of Structure from Motion (SfM) results to train a NeRF-based model to generate TDOM from satellite images using true-ortho-volume rendering.

\begin{figure*}[htbp]
\centering
\includegraphics[width=0.99\textwidth]{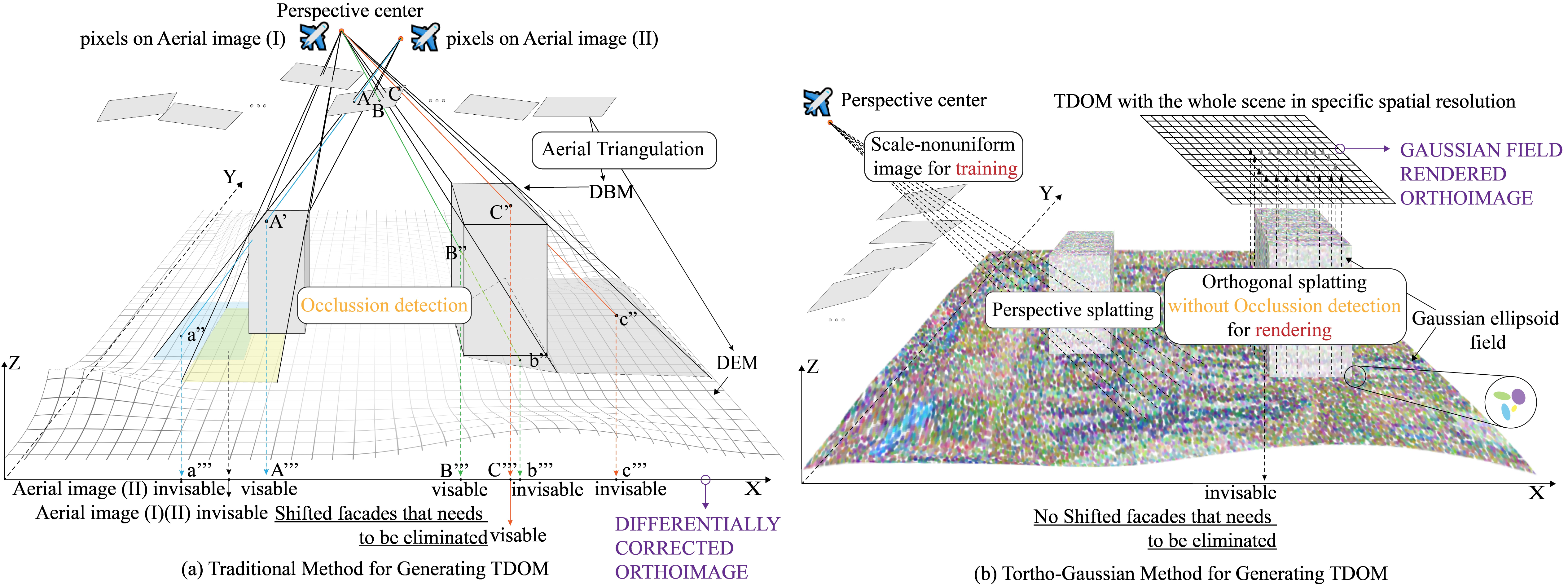}
\caption{Toy examples for generating TDOM. (a) The traditional photogrammetric TDOM generation from multi-view aerial images, incorporating an occlusion detection. Two images I and II are captured from two perceptive centers, with projection lines illustrating occlusion relationships from various viewpoints. In this solution, the DBM (Digital Building Model) and DEM (Digital Elevation Model) are required as input. These assist in occlusion detection, ensuring that only visible building surfaces are displayed, while eliminating façade shifts caused by viewpoint changes. (b) The proposed Tortho-Gaussian, generating a seamless, complete TDOM without the need for mosaicking. Scale-nonuniform and original images are used for 3DGS training. Then, an orthographic projection of the building cluster is performed along the z-axis, bypassing occlusion detection. The Gaussian ellipsoid field is rendered at a selected spatial resolution to produce a true orthophoto of the entire scene. This approach eliminates the need for post-processing steps such as image mosaicking and radiometric/color corrections, offering a streamlined and efficient solution.}
\label{figprinciple}
\end{figure*}

Notwithstanding some advantages of novel view rendering, NeRF and its variants \cite{variants1barron2021mipnerf,variants2liu2020neural,variants3yu2021plenoctrees,20fridovich2022plenoxels} are limited by the time-consuming training process and the high demands for real-time rendering. However, since the last year, 3D Gaussian Splatting (3DGS) has emerged as a promising alternative, offering efficient training and real-time rendering capabilities exceeding 100 frames per second (FPS), thereby becoming a research hotspot. As Fig. \ref{figprinciple} illustrates, 3DGS employs a set of Gaussian Kernels (or ellipsoids) to explicitly represent 3D scene information, which can be reprojected onto a 2D image plane (known as “splatting”). The inherent rasterizer, implemented in CUDA, utilizes parallel-processing tiles to accelerate training and rendering. Consequently, it makes sense to apply 3DGS in the production of TDOMs. However, two significant challenges must be addressed: first, the scalability for large scenes. The number of 3D Gaussian kernels is bounded by the video memory, for example, a 24GB GPU can be used to optimize around 10 million Gaussians, while the small Garden scene of less than 100m2 in the Mip-NeRF360 \cite{barron2022mipnerf360} already needs about 5.8 million 3D Gaussians for rendering. Thus, scalability must be resolved to generate TDOM for large area; second, achieving high fidelity across various terrestrial scenarios. 3DGS often experiences blurring and aliasing when handling strong reflections and slender structures in aerial imagery, such as lakes or power lines.

In this work, we explore the potential of 3D Gaussian Splatting (3DGS) for generating True Digital Orthophoto Maps (TDOMs) through a novel method termed TOrtho-Gaussian. As Fig. \ref{figwhole} shows, we propose the first orthogonal splatting technique for rendering scale-uniform images, aka TDOMs, diverging from conventional perspective splatting. Additionally, inspired by VastGaussian \cite{61lin2024vastgaussian}, we adopt a divide-and-conquer strategy for large-scale scene 3DGS optimization, enabling the rendering of true orthophotos via orthogonal splatting. To achieve a high-fidelity TDOM for challenging scenarios, we introduce a plug-and-play Fully Anisotropic Gaussian Kernel (FAGK), which integrates transparency, scaling, and rotation parameters into spherical harmonic representations. This approach employs dense sampling to significantly enhance the depiction of highly reflective surfaces and slender structures. Our main contributions are as threefold:

\begin{enumerate}
    \item The first attempt to generate TDOM via 3DGS, instead of the perspective splatting in the vanilla 3DGS, presenting a novel orthogonal splatting method which can geometrically elegant bypass the requirement of explicit DSM and occlusion detection.
    \item The application of a divide-and-conquer strategy to extend scalability for generating TDOMs in large-scale scenes, which can improve both the time efficiency pf 3DGS optimization and the video memory usage.
    \item The enhancement of 3D Gaussian representation by Fully Anisotropic Gaussian Kernel, incorporating transparency, scaling, and rotation parameters into spherical harmonic representations.    
\end{enumerate}

\section{Related works}
In this section, some previous relevant works are reviewed including the key challenge for generating TDOM – occlusion detection, learning-based and differentiable Rendering-Based TDOM generation methods.

\subsection{Occlusion Detection-Based TDOM}
To date, ample endeavors have been dedicated to TDOM generation using traditional methods. However, many of these works continue to experience challenges such as misalignment, ghosting, and repeated mapping \cite{08zhou2005comprehensive,18shin2020true,19shin2021true,25sheng2007minimising,33slonecker2009automated,34gilani2016automatic}. These issues often stem from inaccuracy in DSM, but they are primarily exacerbated by insufficient occlusion detection during the digital differential rectification process. To address occlusion detection, various strategies have been developed, including those Z-buffering-based \cite{04amhar1998generation,05rau2000hidden,06nielsen2004true,07uchidatriangle,08zhou2005comprehensive,09habib2006true,10bang2007comparative,11chen2007occlusion,12habib2007new,13antequera2008,14bang2010new}, angles \cite{12habib2007new,22sheng2003true}, heights \cite{08zhou2005comprehensive,09habib2006true,10bang2007comparative,23wang2009new,24oliveira2013occlusion}, vector polygons \cite{08zhou2005comprehensive,15kuzmin2004polygon,25sheng2007minimising,26zhongcheng2010,27xie2010study,28zhong2010vector,29deng2017overall}, texture synthesis \cite{16wang2004method,17zhou2016occlusion}, and object-oriented methods \cite{17zhou2016occlusion,30zitnick2000cooperative,31zhang2012true,32hu2016true}.

The Z-buffer method is one of the earliest techniques introduced for generating orthophoto corrections based on Digital Models (DM) and Digital Building Models (DBM). The basic insight is that two orthophotos are first independently generated and subsequently merged into a single image. The visibility of 3D points is determined by resolving depth conflicts among points illuminated by the same light source, using a perspective projection center \cite{03li2020review,32hu2016true,35haggag2018towards}. This method effectively manages competition among multiple object points along the same projection ray, designating the closest point as visible while marking the others as occluded. Numerous variants of the Z-buffer method have been proposed to enhance its performance \cite{11chen2007occlusion,12habib2007new,13antequera2008,14bang2010new,15kuzmin2004polygon,16wang2004method,17zhou2016occlusion,18shin2020true,19shin2021true}. However, the Z-buffer method is sensitive to the ground sampling distance (GSD) of DSMs and is prone to the M-Portion problem when reconstructing elongated linear structures \cite{03li2020review}. Consequently, many studies \cite{05rau2000hidden,06nielsen2004true,07uchidatriangle,08zhou2005comprehensive,09habib2006true,10bang2007comparative,11chen2007occlusion,12habib2007new,13antequera2008,14bang2010new} have focused on optimizing this method to address occlusion and pseudo-visibility issues.

To further mitigate artifacts, false occlusions, and pseudo-visibility, angle-based methodologies assess occlusion by evaluating the angular relationships between the camera rays of occluded and non-occluded regions in relation to vertical lines \cite{12habib2007new}. For example, Sheng et al. \cite{22sheng2003true}  develop a method for generating angle-based orthophotos of forest scenes using a Canopy Surface Model (CSM), Habib et al.  \cite{12habib2007new}  introduce an occlusion detection technique that calculates off-nadir angles between the lines of the perspective centers and the DSM pixels. Height-based methods have also been proposed, such as Habib et al.'s \cite{09habib2006true} height-based ranking approach and Bang \& Kim's \cite{14bang2010new} height-based ray tracing method. These techniques detect occlusion by utilizing elevation information, identifying occluded regions by comparing building heights with the height of light rays along radial and helical paths \cite{12habib2007new}.

Vector-based methods, such as the one proposed by Sheng et al. \cite{25sheng2007minimising} , aim to eliminate artifacts generated by the Z-buffer technique via treating each pixel in the vector domain as a block rather than a single point. Additionally, Zhong et al. \cite{26zhongcheng2010} introduced an occlusion detection method using polygon-based inversion imaging, leveraging the principle that building polygons do not occlude one another. In addition, texture synthesis methods have been utilized to compensate for missing image information, as demonstrated by Wang et al. \cite{16wang2004method}, who employed a total variation model combined with texture matching. Wang et al. \cite{36wang2018true} further introduced a line segment matching approach, where 2D line segments are extracted and matched, followed by 3D segment reconstruction, resulting in a highly accurate triangulated irregular network (TIN) model before TDOM generation. Li et al. \cite{37li2019fusion} proposed a fusion algorithm based on the pulse-coupled neural network (PCNN) model. Lastly, Zhou et al. \cite{17zhou2016occlusion} developed an object-oriented model that addresses ghosting and occlusion artifacts through a seed growth method to detect occlusions in ghost images. Hu et al. \cite{32hu2016true} proposed a segmentation-based strategy for occlusion compensation, evaluating the simplicity of each segment's cost rather than relying on pixel-level quality assessments.

\subsection{Deep Learning-Based TDOM}
In recent years, various efforts have focused on leveraging deep learning techniques to generate TDOM. Ebrahimikia and Hosseininaveh \cite{46ebrahimikia2022true} propose a solution that improves the quality of DSM through pre-trained deep learning network.  Their approach employs a 2D edge detector to identify building edges in imagery, which are then used to estimate 3D edges, thereby refining the DSM with explicit 3D edge points derived from the point cloud. Subsequently, Ebrahimikia et al.  \cite{47ebrahimikia2024orthophoto} later extend this approach by presenting the Urban-SnowflakeNet, aimed at completing building point clouds from  photogrammetric processing. However, these approaches are limited to structural buildings, lacks generalizability, and still fails to address occlusion detection. Shin et al. \cite{18shin2020true,19shin2021true} enhance the quality of true orthophotos by employing the Pix2Pix model within a Generative Adversarial Network (GAN). Their approach primarily leverages LiDAR data to generate orthophotos that are free from projection geometry occlusions. Nevertheless, their methods heavily depend on the quality of the preprocessed LiDAR intensity.

\begin{figure*}[htbp]
\centering
\includegraphics[width=1.0\textwidth]{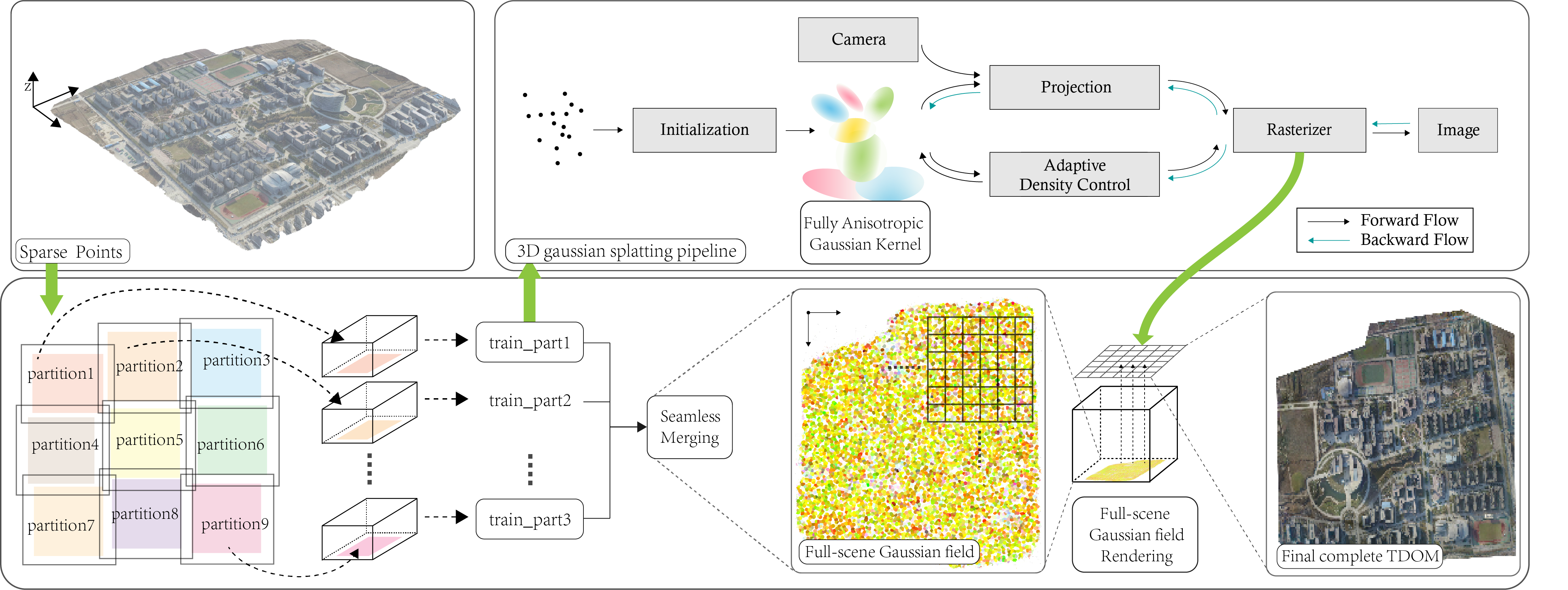}
\caption{Workflow of the proposed Tortho-Gaussian. First, it begins by aligning the sparse point cloud to the \textit{x}- and \textit{y}- axes. The scene is then partitioned into smaller regions using a divide-and-conquer strategy. For each partition, the original 3D Gaussian Splatting (3DGS) framework is employed to train the corresponding Gaussian field with the enhanced Fully Anisotropic Gaussian Kernel (FAGK). The trained Gaussian fields are seamlessly merged into a unified field to represent the entire scene, eliminating the need for TDOM tiling and subsequent color and brightness balancing steps. Next, the camera position is selected, and 3D Gaussian ellipsoids are splatted orthogonally, followed by pixel rasterization to compute the color for each pixel on the TDOM of the complete scene.}
\label{figwhole}
\end{figure*}

\subsection{Differentiable Rendering-Based TDOM}
The rapid advancements in NeRF have profoundly transformed the 2D image reconstruction methods. However, to the best of our knowledge, it is two years later than the NeRF \cite{mildenhall2021nerf} work was proposed that implicit methods began to be leveraged for the generation of true digital orthophotos. Lv et al. \cite{48lv2024fast} are the first to employ implicit neural representations for the rapid generation of digital orthophotos, exploring the potential of implicit reconstruction techniques utilizing the speed-optimized Instant NGP as a baseline. This approach holds the potential to fundamentally alter traditional orthophoto generation practices. Expanding on this groundwork, Chen et al. \cite{49chen2024ortho} employed the speed-optimized Plenoxels \cite{20fridovich2022plenoxels} during the rendering phase, using orthogonal projection to directly produce high-quality TDOM. This method effectively circumvents the intricate challenges associated with visibility analysis and texture compensation that are commonly encountered in conventional approaches, while also mitigating issues related to terrain edge distortion. Later, Qu et al. \cite{62liu2025citygaussian} further extend this idea by inputting satellite images for generating more large-arear TDOM.

Recently, 3DGS, as a differentiable rendering method that incorporates explicit spatial structural information, has started to supplant tasks accomplished by NeRF \cite{50chen2024survey,51dalal2024gaussian,52fei20243d,53wu2024recent}. This shift is attributed to 3DGS’s exceptional speed and high-detail novel view synthesis quality \cite{54cotton2024dynamic,55wu20244d}. Its flexible and adaptive 3D object representation overcomes the limitations of traditional volumetric rendering methods \cite{52fei20243d}. Therefore, this study seeks to explore the feasibility and potential of 3DGS for generating TDOM.

\section{Methodology}
This section provides detailed explanations of our Tortho-Gaussian on generating TDOM, which contains five technical parts, i.e., preliminaries of 3GDS (\ref{prel}), orthographic splatting of 3DGS (\ref{TDOM}), TDOM generation based on orthographic splatting (\ref{Ortho Splatting}), fully anisotropic Gaussian Kernels (\ref{FAGK}) and the divide-and-conquer strategy (\ref{Divide and Conquer strategy}) for large-scale area. More specifically, as Fig. \ref{figwhole} shows, after structure from motion, our Tortho-Gaussian first divide the whole scene into sub-regions, which are optimized by the 3DGS with an improved fully anisotropic Gaussian Kernel. After the optimization, we algin the 3DGS field such that the scene is parallel to the xy-plane. Then, the centroid of all camera are explored to set the spatial resolution, and performing orthogonal splatting along the z-axis to produce a true orthophoto of the complete scene, without the need for image stitching or color correction.

\subsection{Preliminaries} \label{prel}
Following portion of \cite{56kerbl20233d}, to make this paper more self-contained, we next outline some basics of 3D gaussian splatting. The 3DGS represents the scene using a series of Gaussian ellipsoid kernels that are closely aligned with the 3D scene's structures. Each Gaussian kernal is defined by a set of attributes, including the position (mean) $\mu $, anisotropic covariance $\Sigma$, opacity $\alpha $, and color $\text{c}$ that is formulated via spherical harmonics. During the rendering stage, the position and covariance attributes of all Gaussians in the 3D scene are reprojected onto the image plane (namely Splatting), thereby forming 2D Gaussians. The reprojection of 3D Gaussians onto specific image tiles is determined based on the position and radius of these 2D Gaussians. The rendered image is then generated using a volumetric rendering technique combined with alpha blending. Ultimately, the Gaussian kernels are optimized based on the discrepancy between the rendered image and the input image. A Gaussian kernel $G_{\Sigma}(\mathbf{x})$, centered at $\mu $, with a 3D covariance matrix given by $\Sigma$ is represented via following formula.
\begin{equation}
\label{eq1}
G_{\Sigma}(\mathbf{x}-\mu)=\frac{1}{(2\pi)^{3/2}\mid\Sigma\mid^{1/2}}e^{-\frac{1}{2}(\mathbf{x}-\mu)^{T}\Sigma^{-1}(\mathbf{x}-\mu)}
\end{equation}
Where x and $\mu $ represent column vectors, specifically $[x,y,z]^{T}$ and $[\mu_x,\mu_v,\mu_z]^T$, respectively. The term $\Sigma$ denotes a symmetric 3 × 3 matrix. In order to maintain its positive semi-definite property, the covariance matrix is further parameterized by a scaling matrix $\text{S}$ and a rotation matrix $\text{R}$:
\begin{equation}
\Sigma=RSS^{T}R^{T}
\end{equation}
To perform the splatting, the 3D Gaussian kernel is projected onto a 2D Gaussian ellipse. The corresponding 2D covariance matrix $\Sigma^{\prime}$ is calculated by the following formula:
\begin{equation}
\Sigma^{^{\prime}}=JW\Sigma W^TJ^T
\end{equation}
 $\text{W}$ denotes the viewpoint transformation, and $\Sigma$ represents the 3D covariance matrix, while  $\text{J}$ denotes the Jacobian matrix associated with the projective transformation within its affine approximation \cite{56kerbl20233d,57zwicker2001ewa}.

For a specific pixel, the corresponding depths of all intersecting Gaussians can be derived with the viewing transformation $\text{W}$, which are employed to sort these intersected Gaussians \textbf{\textit{$\mathcal{N}$}}. Subsequently, the final color of the pixel is calculated using alpha blending:
\begin{equation}
C=\sum_{n=1}^{|\mathcal{N}|}c_n\alpha_n^{'}\prod_{j=1}^{n-1}\left(1-\alpha_j^{'}\right)
\end{equation}
$c_{n}$  denotes the predicted color. The final splatting opacity  $\alpha_{n}^{'}$  is obtained by multiplying the predicted opacity $\alpha_{_n}$  with the splatted 2D Gaussian, as defined below:

\begin{equation}
\alpha_n^{\prime} = \alpha_n e^{ -\frac{1}{2} (x^{\prime} - \mu_n^{\prime})^{\mathrm{T}} \Sigma_n^{\prime -1} (x^{\prime} - \mu_n^{\prime}) }
\end{equation}
 \text{x'} and $\mu_{n}^{'}$ are coordinates defined in the 2D image plane. .

\subsection{Orthogonal Splatting of 3DGS}\label{TDOM}
The 3D Gaussian Splatting technique offers significant potential for accurate scene representation and rendering, similar to the advancements in TDOM generation seen with NeRF \cite{48lv2024fast,49chen2024ortho}. Based on the vanilla 3DGS that employs perspective splatting, we introduce orthogonal splatting which mainly includes the projection of the mean and variance of 3D Gaussian.
\subsubsection{Projection of 3D Gaussian - Mean}
The Gaussian mean represents the position coordinates of the Gaussian kernel, which is capable of undergoing a non-affine transformation through perspective projection. The corresponding transformation matrix for this projection is given by:
\begin{equation}
\label{eqp1}
P=\begin{pmatrix}\frac{2 z_{\mathrm{n}}}{r-l}&0&\frac{r+l}{r-l}&0\\0&\frac{2 z_{\mathrm{n}}}{t-b}&\frac{t+b}{t-b}&0\\0&0&-\frac{z_{\mathrm{f}}+z_{\mathrm{n}}}{z_{\mathrm{f}}-z_{\mathrm{n}}}&-\frac{2 z_{\mathrm{f}} z_{\mathrm{n}}}{z_{\mathrm{f}}-z_{\mathrm{near}}}\\0&0&-1&0\end{pmatrix}
\end{equation}
where
\begin{equation}
r=\tan\left(\frac{\theta_{x}}{2}\right)\cdot z_{\mathrm{n}}\text{,  }l=-r
\end{equation}
\begin{equation}
t=\tan\left(\frac{\theta_{y}}{2}\right)\cdot z_{\mathrm{n}}\text{,  }b=-t
\end{equation}
$z_{\mathrm{n}}$ and $z_{\mathrm{f}}$ represent the near and far clipping planes, respectively, while $\theta_{x}$ and $\theta_{y}$ are the horizontal and vertical fields of view. \textit{l}, \textit{r}, \textit{b} and \textit{t} denote the left, right, top, and bottom boundaries of the viewing frustum. The Gaussian kernel is projected onto the 2D image plane through this perspective projection.

In order to make an orthographic projection for the mean value, equation (\ref{eqp1}) can be replaced by the following formula:
\begin{equation}
\label{eqortho}
P_o=\begin{pmatrix}\frac{2}{r-l}&0&0&-\frac{r+l}{r-l}\\0&\frac{2}{t-b}&0&-\frac{t+b}{t-b}\\0&0&-\frac{2}{z_{\mathrm{f}}-z_{\mathrm{n}}}&-\frac{z_{\mathrm{f}}+z_{\mathrm{n}}}{z_{\mathrm{f}}-z_{\mathrm{n}}}\\0&0&0&1\end{pmatrix}
\end{equation}

Via equation (\ref{eqortho}), the center of the Gaussian sphere is orthographically splatted into a corresponding 2D Gaussian.

\subsubsection{Projection of 3D Gaussian Axes (Covariance)}

\cite{58zwicker2002ewa} proved that covariance matrix can be represented using rotation and scaling matrices. In the context of projective transformations, the expected perspective projection matrix is given by formula (\ref{eqp1}). It is evident that this transformation is nonlinear and non-affine. To approximate the affine transformation of the Gaussian ellipsoid, a corresponding Jacobian matrix is employed for local linear approximation:
\begin{equation}
J=\begin{pmatrix}\frac{focal_x}{t_z}&0&-\frac{focal_x\cdot t_x}{t_z^2}\\0&\frac{focal_y}{t_z}&-\frac{focal_y\cdot t_y}{t_z^2}\\0&0&0\end{pmatrix}
\end{equation}
where $focal_{x}$ and $focal_{y}$ represents the focal lengths of the camera along the x and y axes. $(t_x,t_y,t_z)$ is the coordinates of a 3D point in camera space and $(\nu_x,\nu_y,\nu_z,1)$ is the corresponding homogeneous coordinate.  Then, the Jacobian matrix $J_{\mathbf{o}}$ corresponding to the orthogonal matrix $P_{o}$ can be derived by formula (\ref{eqv'}).

\begin{equation}
\label{eqv'}
\nu'=P_o\begin{pmatrix}\nu_x\\\nu_y\\\nu_z\\1\end{pmatrix}=\begin{pmatrix}\frac{2}{r-l}\nu_x+\frac{r+l}{r-l}\\\frac{2}{t-b}\nu_y+\frac{t+b}{t-b}\\-\frac{2}{z_\text{f}-z_\text{n}}\nu_z+\frac{z_\text{f}+z_\text{n}}{z_\text{f}-z_\text{n}}\\1\end{pmatrix}
\end{equation}
Differentiating this equation, we can obtain: 
\begin{equation}
J_o=\begin{pmatrix}\frac{\partial\nu_x'}{\partial\nu_x}&\frac{\partial\nu_x'}{\partial\nu_y}&\frac{\partial\nu_x'}{\partial\nu_z}\\\frac{\partial\nu_y'}{\partial\nu_x}&\frac{\partial\nu_y'}{\partial\nu_y}&\frac{\partial\nu_y'}{\partial\nu_z}\\0&0&0\end{pmatrix}=\begin{pmatrix}\frac{2}{r-l}&0&0\\0&\frac{2}{t-b}&0\\0&0&0\end{pmatrix}
\end{equation}

To accurately orthogonally splat a 3D Gaussian onto a referenced 2D plane, both the Gaussian mean and variance have to undergo appropriate orthographic projection to ensure correct coverage of the corresponding tiles and pixels. Subsequently, this allows each pixel to be encoded as related 2D Gaussian in an orthographic manner, which are then rendered using $\alpha$-blending based on the depth order.

\subsection{TDOM generation based on orthographic splatting}\label{Ortho Splatting}
To generate the TDOM of the whole scene, we have to perform an accurate orthographic projection for the entire scene, while avoiding the need for image mosaicking. First, the target spatial resolution $s_{x}$ and $s_{y}$ for each pixel on the TDOM should be set. During orthographic splatting, the grid points (as shown in Fig. \ref{figprinciple} and \ref{figwhole}), formed by the pre-set spatial resolution, are applied as referenced coordinates for rasterization.  The 3D Gaussians that locate within one specific grid are employed to render the color information of corresponding TDOM pixel. The coordinates of each TDOM pixel can be defined as follows:
\begin{equation}
X=\left\{\bar{X}+s_x\cdot(i-(W/2)+\delta_x)\mid i=0,1,\ldots,W\right\}
\end{equation}
\begin{equation}
Y=\begin{Bmatrix}\bar{Y}+s_y\cdot(j-(H/2)+\delta_y)\mid j=0,1,\ldots,H\end{Bmatrix}
\end{equation}

where, \textit{W} and \textit{H} is the width and Height of the TDOM which is up to the spatial resolution, $\delta_x$ and $\delta_y$ is constant value (equal to 1/2 spatial resolution) to make projection ray stay at the center of the pixel. For the position coordinates of $\text{\textit{N}}$ cameras, denoted as $\{(x_i,y_i,z_i)\mid i=1,2,\ldots,N\}$, the center of the orthographic image, denoted as $(\bar{X},\bar{Y})$ can be calculated using the centroid of the cameras' positions in the \textit{x}- and \textit{y}- directions:
\begin{equation}
\bar{X}=\left\{\frac{1}{N}\sum_{i=1}^Nx_i\right\}
\end{equation}
\begin{equation}
\bar{Y}=\left\{\frac{1}{N}\sum_{i=1}^{N}y_i\right\}
\end{equation}

Finally, for each pixel, we orthographically splat the corresponding 3D Gaussians into 2D Gaussians, which are then $\alpha$-blended based on the depth sort for esimating the color. The complete TDOM is obtained by running all pixels as above.
\subsection{Fully Anisotropic Gaussian Kernel}\label{FAGK}
Inspired by MLP-based Gaussian kernels \cite{59lu2024scaffold}, we develop a novel fully anisotropic Gaussian kernel that incorporates spherical harmonic (SH) coefficients for all corresponding properties, including color, opacity, rotation, and scaling. This representation enables the Gaussian kernel at a given position to exhibit distinct colors $\textit{c}$, opacities $\alpha $, rotations $\textit{r}$ , and scales $\textit{s}$ when observed from different directions, allowing our approach to achieve comparable performance to neural Gaussians \cite{59lu2024scaffold}.

Specifically, the direction between the Gaussian kernel and the camera center is first calculated resulting in direction-dependent Gaussian properties. The complexity of the spherical harmonics depends on the degree of the SH coefficients. In this work, similar to 3DGS, the coefficients up to the third order are applied, and upgrade the order every 1000 iterations. For each channel of Gaussian properties, the spherical harmonics are formulated as:
\begin{equation}
A_l^m(\theta,\varphi)=\begin{cases}\sqrt{2}K_l^mcos(m\varphi)P_l^m(cos\theta)&m>0\\\sqrt{2}K_l^msin(-m\varphi)P_l^{-m}(cos\theta)&m<0\\K_l^0P_l^0(cos\theta)&m=0\end{cases}
\end{equation}
where
\begin{equation}
\begin{gathered}
P_{n}\left(x\right)=\frac{1}{2^{n}\cdot n!}\frac{d^{n}}{dx^{n}}\Big[\left(x^{2}-1\right)^{n}\Big] \\
P_{l}^{m}=(-1)^{m}\left(1-x^{2}\right)^{\frac{m}{2}}\frac{d^{m}}{dx^{m}}(P_{l}(x)) \\
K_l^m=\sqrt{\frac{(2l+1)(l-\mid m\mid)!}{4\pi(l+\mid m\mid)!}} 
\end{gathered}
\end{equation}

$K_l^m$ is the normalization constant, with  $\textit{l}$ as the degree and $\textit{m}$  as the order of the spherical harmonics. $A_{l}^{m}(\theta,\phi)$ is the general form of the spherical harmonics. $P_{l}^{m}(x)$ is the 
associated Legendre polynomial. $\theta $  and $\varphi $ are the polar and azimuthal angles in spherical coordinates, respectively .

\subsection{Divide and Conquer strategy}\label{Divide and Conquer strategy}
The vanilla 3DGS uses explicit 3D Gaussian ellipsoids as primitives to represent a 3D scene. However, it is limited in handling only a finite number of 3D Gaussian ellipsoids, which is typically feasible for small-scale objects \cite{60chen2024dogaussian,61lin2024vastgaussian,62liu2025citygaussian,63xie2024gaussiancity}. As the scale of the scene increases, the demand for GPU memory (VRAM) grows substantially. For example, given a 24GB RTX 3090 GPU, it runs into memory shortage when the number of Gaussians exceeds approximately 10 million. Reconstructing large-scale scenes typically requires significantly higher number of Gaussians, further exacerbating this issue \cite{63xie2024gaussiancity}. Additionally, the increasement of Gaussians also leads to a considerable slowdown in the depth sorting process prior to rendering. 
\begin{figure}[htbp]
\centering
\includegraphics[width=0.48\textwidth]{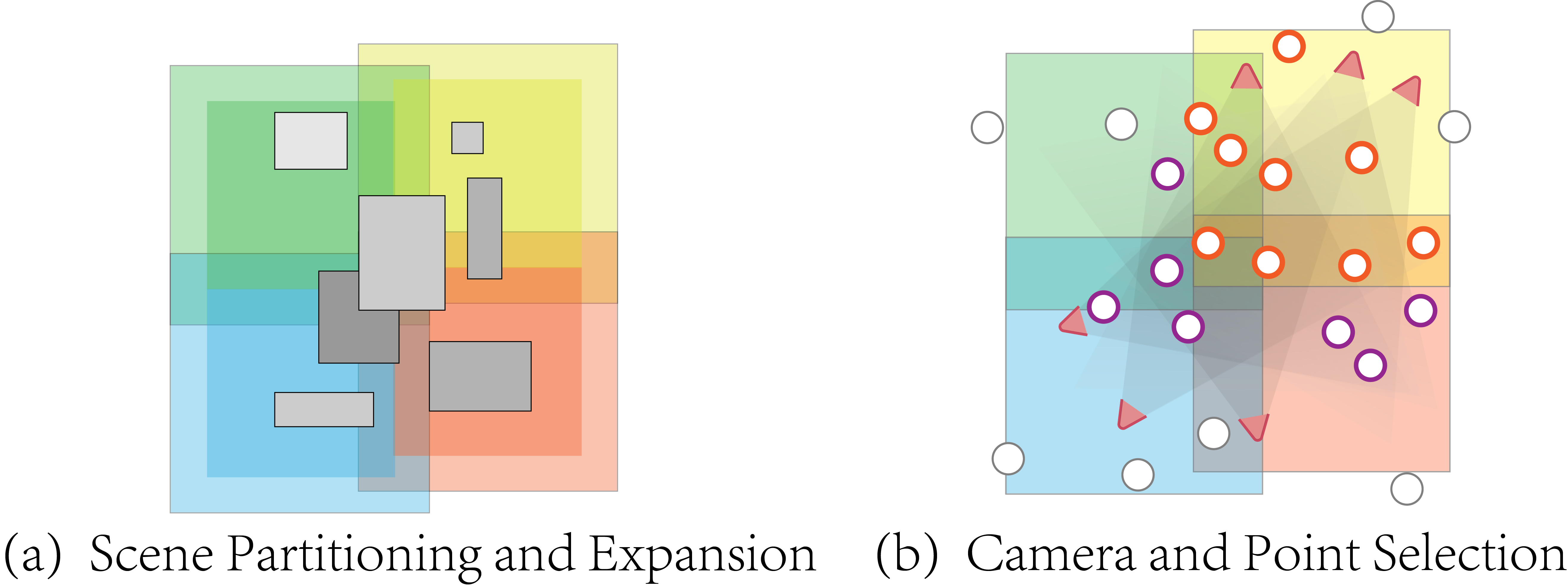}
\caption{Divide-and-Conquer strategy. The dark rectangles are the initial divisions of the scene, the light-colored rectangles indicate the extended regions, and the orange and purple dots represent cameras selected by the local region and the external region,
respectively.}
\label{figpartion}
\end{figure}
 \begin{figure*}[htbp]
\centering
\includegraphics[width=1.05\textwidth]{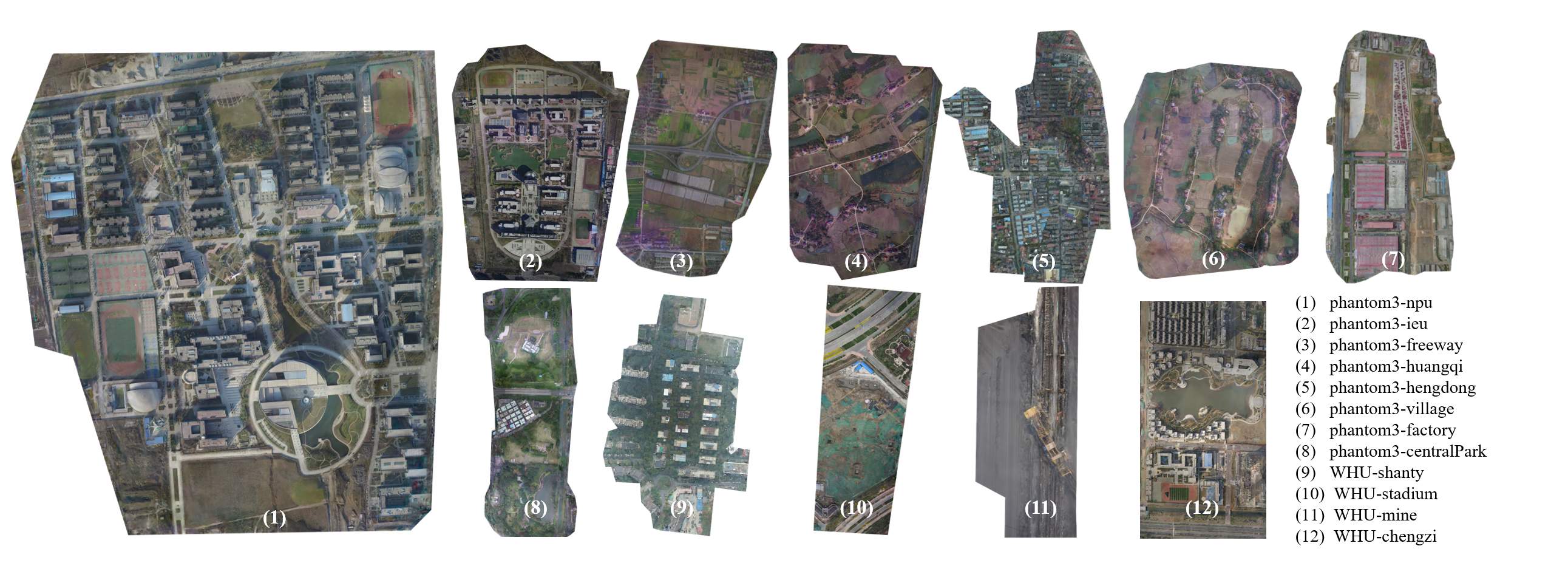}
\caption{The complete TDOMs of the NPU-DroneMap and WHU dataset using our proposed Tortho-Gaussian.}
\label{figstich}
\end{figure*}

To address this issue, analogous to \cite{61lin2024vastgaussian}, we adopted a divide-and-conquer strategy, introducing a progressive partition approach. Basically, the key insight is that the scene is divided into multiple overlapping rectangular regions for optimization with limited VRAM, reducing both training time and memory requirement. Fig. \ref{figpartion} illustrates the strategy we employ, and four steps are performed for the division: 

\subsubsection{Scene Partitioning Based on Camera Positions}
In general, the input images are distributed across the scene to facilitate comprehensive reconstruction. Therefore, dividing the scene based on the distribution of camera centers makes sense. The camera centers are projected onto the ground plane. An   $m\times n$  grid is adopted and ensures each cell contains a similar number of training images, thereby balancing the optimization procedure across different cells. As \cite{61lin2024vastgaussian}, the views $\textit{V}$ on ground plane is divided into \textit{\textit{$ m$ }} sections, each containing approximately $\left|\mathbf{V}\right|/m$   views. These sections are further subdivided into  \textit{$ n$ }cells, with around $|\mathbf{V}|/(m\times n)$ views per cell. 
\subsubsection{Sub-division expansion}
To guarantee the consistency between adjacent subdivisions, it is necessary to expand the partitioned cells to some degrees. As Fig. \ref{figpartion} (a) illustrates that the light-colored rectangles are from the extended darker ones. For the \textit{j\textsuperscript{th}} cell, defined within a rectangle of size   $\ell_i^h\times\ell_i^w$, we expand along the boundary by 20\%, leading to a larger cell. This expanded boundary ensures more comprehensive coverage which contained extended subsets of camera set $\{\mathbf{V}_{i}\}_{i=1}^{m\times n}$ and point set $\{\mathbf{P}_{i}\}$ for each cell, allowing for improved continuity between adjacent cells, 
\subsubsection{Camera Selection Based on Visibility}
Incorporating more images generally provides additional training samples, thereby enhancing the performance of 3DGS. For each divided cell, Fig. \ref{figpartion} (b) shows the selection of additional cameras based on visibility to enhance reconstruction fidelity. Cameras with visibility values exceeding a predefined threshold \textbf{\textit{t}} are chosen, where visibility is defined as the ratio of the projected area of the divided cell onto an image to the area of the entire image.

\begin{table*}[htbp]
\centering
\caption{Details of our experimental datasets}
\label{table:data_used}
\renewcommand{\arraystretch}{1.2}
\begin{tabular}{ccccccc}
\toprule
\textbf{Dataset} & \textbf{Sequence} & \textbf{Location} & \textbf{H-Max (m)} & \textbf{Area (km²)} & \textbf{Imagenum} & \textbf{Imagesize (pix)} \\
\midrule
\multirow{9}{*}{NPU-DroneMap} & phantom3-hengdong & Hengdong, Hunan & 358.00 & - & 221 & 1920*1080 \\
 & phantom3-huangqi & Hengdong, Hunan & 222.30 & 1.313 & 393 & 1920*1080 \\
 & phantom3-centralPark & Shenzhen, Guangdong & 161.80 & 0.606 & 835 & 1920*1080 \\
 & phantom3-factory & - & 198.72 & 0.912 & 402 & 1920*1080 \\
 & phantom3-freeway & - & 258.30 & 1.457 & 415 & 1920*1080 \\
 & phantom3-village & Hengdong, Hunan & 160.60 & 0.932 & 406 & 1920*1080 \\
 & phantom3-ieu & Zhenzhou, Henan & 282.30 & 1.524 & 467 & 1920*1080 \\
 & phantom3-npu & Xi’an, Shaanxi & 254.50 & 1.598 & 457 & 1920*1080 \\
\midrule
\multirow{3}{*}{WHU} & shanty & Wuhan, Hubei & 120.00 & 0.503 & 67 & 1600*900 \\
 & stadium & Tianjin & 310.00 & - & 230 & 1600*900 \\
 & mine & Rizhao, Shandong & 60.60 & - & 43 & 1600*900 \\
 & chengzi & Suqian, Jiangsu  & 167.35 & - & 819 & 1228*820 \\
\bottomrule
\multicolumn{7}{l}{\textit{'-' denotes missing information.}}
\end{tabular}
\end{table*}

 \subsubsection{Point Selection Based on Coverage}
After adding relevant cameras to the cell's camera set  $\mathbf{V}_{i}$ , all the points observed by these extra selected cameras are added to the point set  $\mathbf{P}_{i}$. This step ensures a better initialization for the optimization of each cell. Proper initialization helps mitigate depth ambiguities, which could otherwise lead to incorrect 3D Gaussian distributions when fitting objects, particularly those located outside the cell. This method ensures accurate 3D Gaussian generation and prevents the formation of floating artifacts. 

\section{Experiments}
In this section, comprehensive experimental results are reported to demonstrate the efficacy of the proposed Tortho-Gaussian. Based on various datasets, both qualitative and quantitative evaluations are conducted via comparison with several state-of-the-art commercial software and ablation studies.
\begin{figure*}[htbp]
\centering
\includegraphics[width=0.85\textwidth]{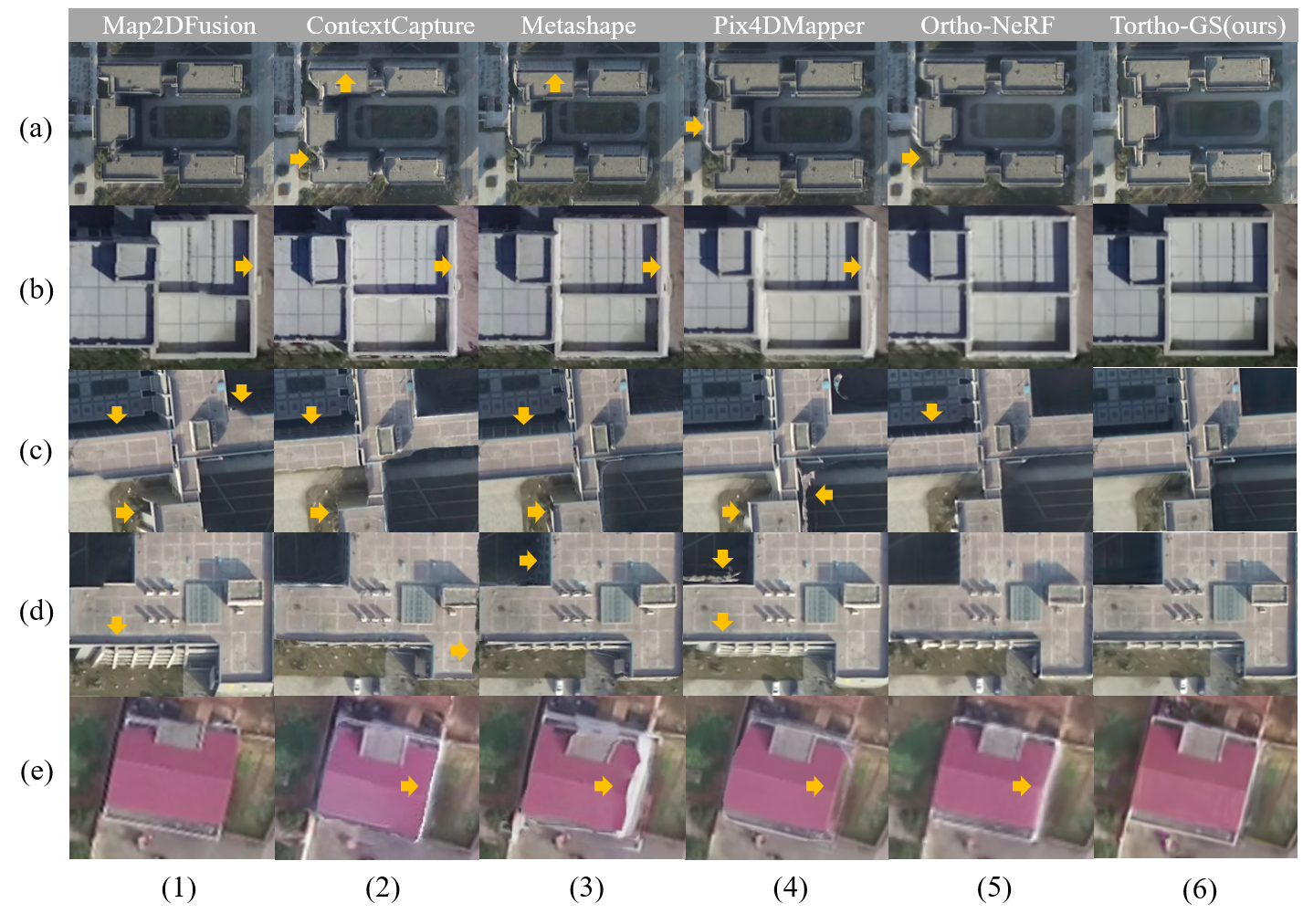}
\caption{Qualitative comparison of TDOMs generated by Map2DFusion, commercial software, Ortho-NeRF and our method on NPU DroneMap dataset. Our method can effectively reconstruct linear building edges and shows superior performance in eliminating distortions of building facades.}
\label{figfruit1}
\end{figure*}
\subsection{Experimental settings}\label{settings}
\textbf{Experiments protocols.} Overall, our experiments mainly contains three parts: first, Qualitative Performance. We compare our method with several state-of-the-art (SOTA) commercial software. Specifically, the quality of the generated TDOM is evaluated by analyzing building edges, facades, slender structures, and regions with weak textures; second, quantitative performance. To assess the precision of the generated TDOM, we first evaluate relative precision using Metashape and Pix4DMapper as benchmarks. Additionally, we provide an overlaid comparison with manually generated CAD maps to quantify the absolute mapping precision; Finally, ablation studies. We conduct extensive ablation studies to explore various aspects of our Tortho-Gaussian method, including spatial resolution, partitioning strategies, and the use of fully anisotropic Gaussian kernels. This comprehensive evaluation highlights the robustness and effectiveness of our approach.

\textbf{Experimental Data.} In our work, we first employ the NPU DroneMap dataset published by Bu et al. \cite{64bu2016map2dfusion}, it was captured images across different areas in China using a custom-built hexacopter equipped with a Phantom3 camera. Additionally, we generate a self-constructed dataset, WHU, including Shanty, Stadium, Mine and chengzi datasets using on Canon EOS 5D Mark III under different flight Height, among which the chengzi dataset contain a high-precision CAD map that is made by manual labeling. These experimental data consists of high-resolution aerial imagery captured in urban, rural, and mixed environments, encompassing various scene elements such as buildings, vegetation, roads, and water bodies.  The overall views of these datasets are shown in Fig. \ref{figstich}, more detailed information can be found in Tab. \ref{table:data_used}.

\begin{figure}[htbp]
\centering
\includegraphics[width=0.45\textwidth]{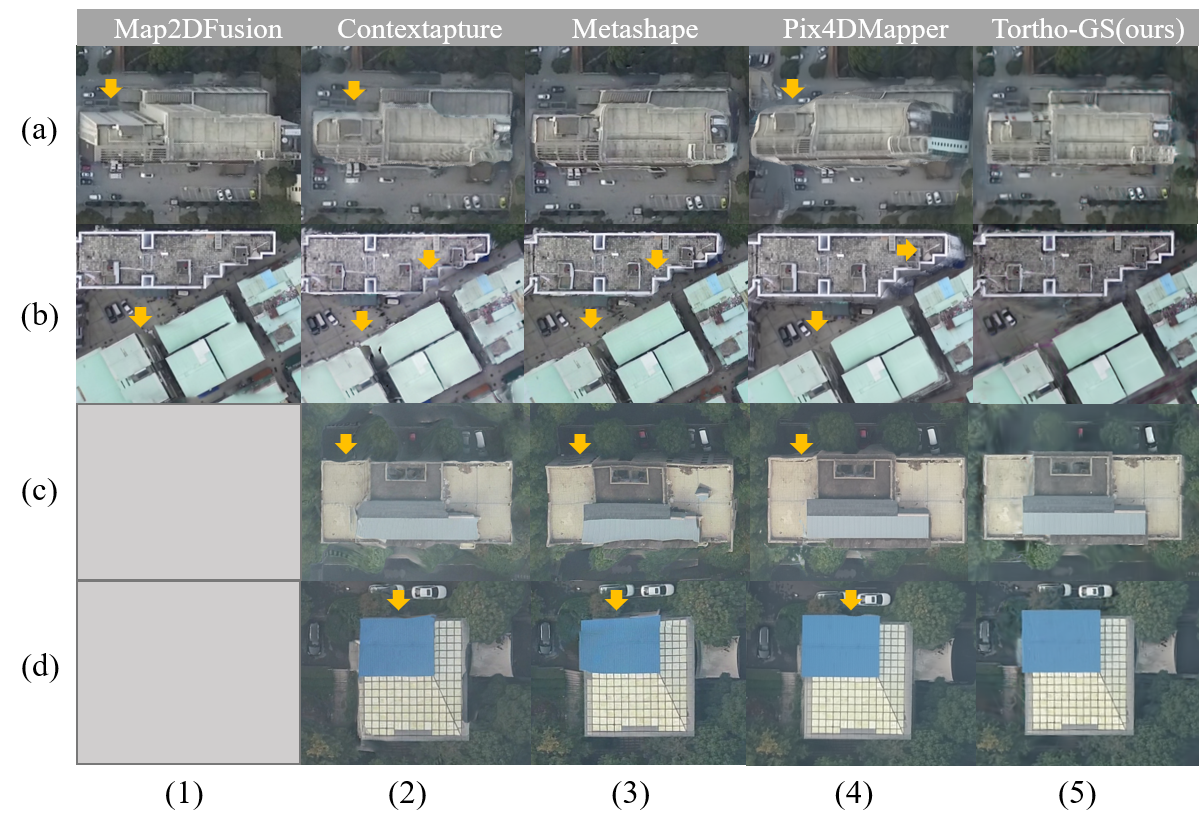}
\caption{Qualitative comparison of TDOM generated by Map2DFusion, commercial software, and our method. The gray block denotes the result is not available due to the practical re-implementation issue.}
\label{figfruit2}
\end{figure}

\textbf{Experimental Details}. Before 3D Gaussian optimization, the sparse point cloud of SfM was first preprocessed by Manhattan alignment, ensuring that the x and y axes of the point cloud were parallel to the boundary frame. Then, the resulting transformation matrix was applied to each sub-regional training. During optimization, for every fixed number of images, one image was selected as a test image. All experiments were conducted on four NVIDIA GeForce RTX 4090 GPUs, with the number of iterations set to 30,000.

\subsection{Qualitative Evaluation}
 Four commercial software (ContextCapture \cite{65bentley_contextcapture}, Metashape \cite{66agisoft_metashape}, Pix4DMapper \cite{67pix4d_mapper}, and Map2DFusion \cite{64bu2016map2dfusion}), incorporated with traditional photogrammetric techniques relying on DSM, are compared on the NPU DroneMap and WHU dataset regarding the quality of TDOM. All these methods use the poses of ContextCapture before conducting the subsequent reconstruction tasks. Two indicators that can significantly reflect the quality of TDOM are visually investigated, i.e., building edges and facades.

\subsubsection{Building Edges}
In general, the edges of buildings on a satisfactory TDOM should follow the real geometry of building without irregular deformation, and the joints between buildings should precisely align.  Fig. \ref{figfruit1} and Fig. \ref{figfruit2} shows the results of TDOM in a region densely packed buildings, and several different methods are compared. As it can be seen from  Fig. \ref{figfruit1}  (e) and Fig. \ref{figfruit2}  (a), the reconstructed edges in Map2DFusion, ContextCapture, Metashape and Pixel4DMapper all exhibit varying degrees of distortion due to the insufficient digital differential correction. Even the Nerf-based method Ortho-NeRF \cite{49chen2024ortho} shows some artifacts that may be influenced by the inherent rendering performance of neural-based implicit 3D representation. On the other hand, our Tortho-Gaussian successfully generates high-fidelity building edges.

To further explore the quality of edge reconstruction, six buildings with prominent straight lines are selected from the NPU Dronemap and WHU datasets for extra assessment. In particular, we applied the Canny edge detection algorithm \cite{68canny1986computational}, using minimum and maximum thresholds, to extract line edge structures from the building outlines\footnote{Empirically, we set the minimum threshold of the Canny algorithm as 50\% of the maximum threshold to detect primary edge features with significant pixel gradients. The dimension of the Sobel operator, or aperture size, was set to 3}. Based on the discrete points extracted on the generated TDOM, the lines are fitted using least-squares in an iterative manner. The points that stay within 2.5 time the standard deviation are considered as inliers and used to reflect quality of line edges obtained by various methods.

\begin{figure}[htbp]
\centering
\includegraphics[width=0.5\textwidth]{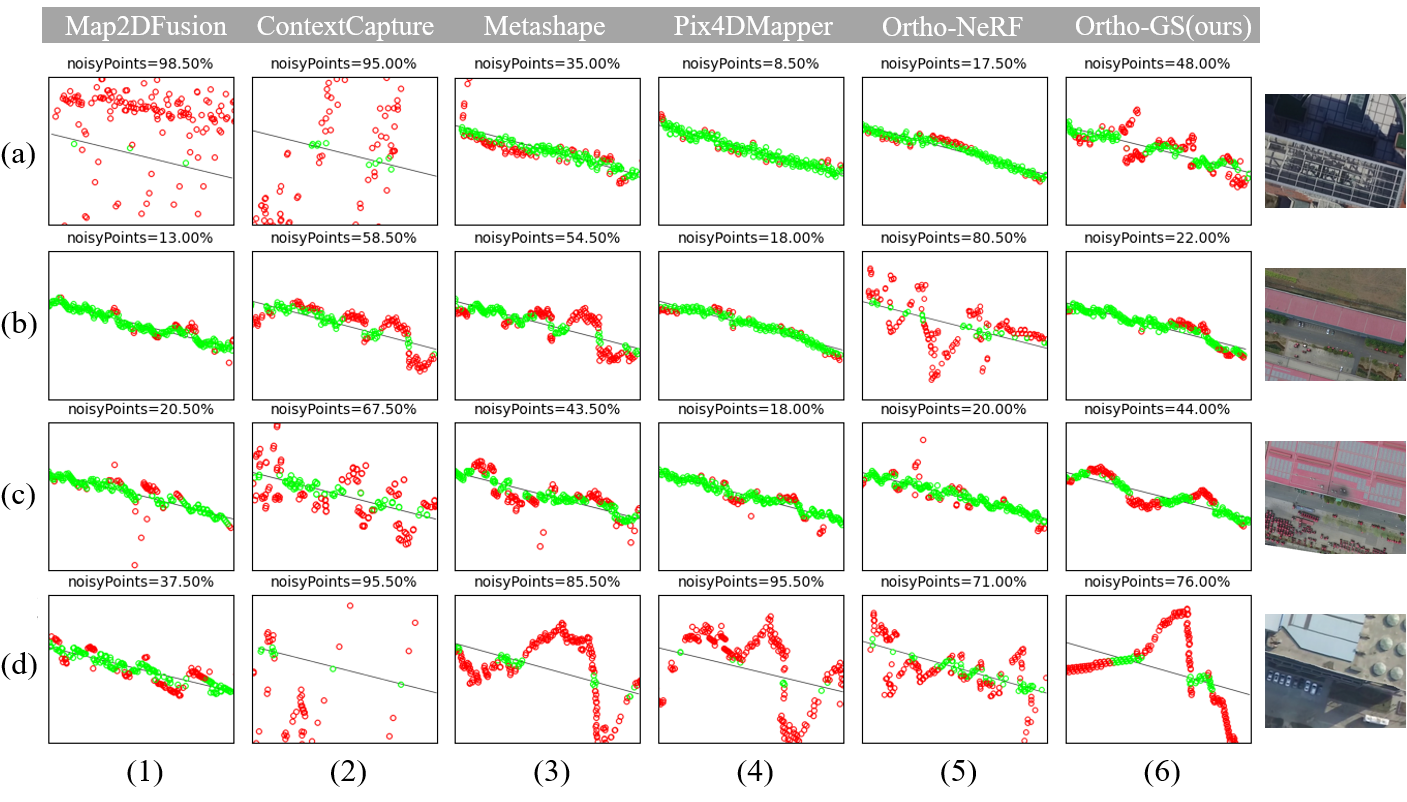}
\caption{Analysis of the building edges on various TDOM generated by Map2DFusion, commercial software, Ortho-NeRF and our method on NPU DroneMap dataset.}
\label{figvalue1}
\end{figure}

\begin{figure}[htbp]
\centering
\includegraphics[width=0.5\textwidth]{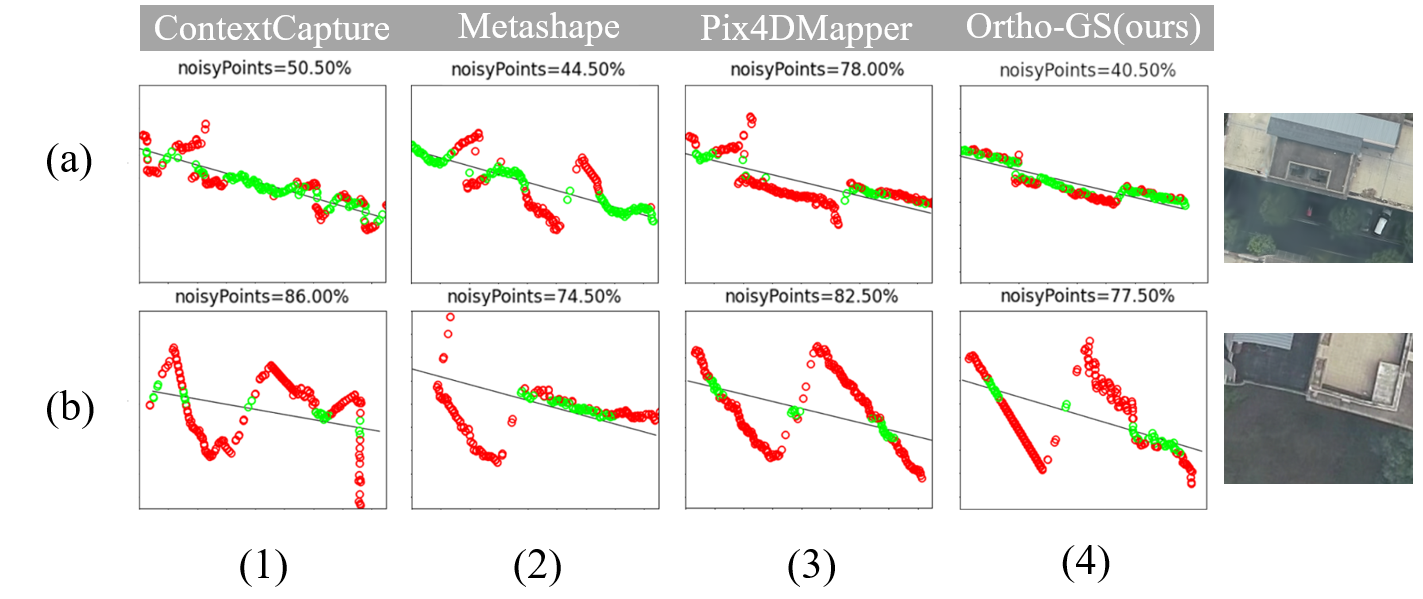}
\caption{Analysis of the building edges on various TDOM generated by commercial software,  and our method on WHU dataset.}
\label{figvalue2}
\end{figure}

\begin{table*}[htbp]
\centering
\caption{Relative precision of Tortho-Gaussian comparing to Metashape and Pix4DMapper}
\label{table:error_comparison}
\renewcommand{\arraystretch}{1.2}
\begin{tabular}{ccccccccc}
\toprule
\textbf{ID} & \textbf{Tortho-Gaussian} & \multicolumn{3}{c}{\textbf{Metashape}} & \multicolumn{3}{c}{\textbf{Pix4DMapper}} \\
\cmidrule(lr){2-2}\cmidrule(lr){3-5} \cmidrule(lr){6-8}
& \textbf{Ratio} & \textbf{Ratio} & \textbf{Relative Error (\%)} & \textbf{Absolute Error}  & \textbf{Ratio} & \textbf{Relative Error (\%)} & \textbf{Absolute Error}  \\
\midrule
1  & 1.41978 & 1.42071 & 0.06578 & 0.000935 & 1.42186 & 0.14623 & 0.001145 \\
2  & 0.54305 & 0.54413 & 0.19844 & 0.001080 & 0.54253 & 0.09648 & 0.001603 \\
3  & 1.01485 & 1.01614 & 0.12680 & 0.001288 & 1.01721 & 0.23185 & 0.001700 \\
4  & 1.22322 & 1.22181 & 0.11548 & 0.001411 & 1.22285 & 0.02993 & 0.001405 \\
5  & 0.81333 & 0.81238 & 0.11662 & 0.000947 & 0.81371 & 0.04705 & 0.001303 \\
6  & 3.04875 & 3.04673 & 0.06639 & 0.002023 & 3.04333 & 0.17822 & 0.005424 \\
7  & 1.04369 & 1.03974 & 0.37979 & 0.003499 & 1.04526 & 0.15076 & 0.001036 \\
8  & 1.21988 & 1.22197 & 0.17061 & 0.001793 & 1.21833 & 0.12474 & 0.001557 \\
\midrule
\textbf{Mean} & - & - & \textbf{0.15486} & \textbf{0.001622} & - & \textbf{0.12591} & \textbf{0.001772} \\
\bottomrule
\end{tabular}
\end{table*}
As shown in Fig. \ref{figvalue1} and Fig. \ref{figvalue2}, comparing to the other methods, in general, our method can reduce the number of noise points in the linear building edges and achieve SOTA performance. Our Tortho-Gaussian is basically on par with other methods, wherein we yield the best results on some tests and others generate the best on other datasets. 

These experimental results demonstrate that our approach can effectively suppresses jagged edges and curvature along building outlines.


\subsubsection{Building Facades}
For true orthophotos, the side facades of building should appear as just continuous boundaries, without breaks or jagged edges.Typically, the extent to which building facades can be fully resolved reflects the effectiveness of occlusion detection in the true orthophoto generation process. Our method effectively eliminates almost all facade occlusions. This can be clearly observed in Fig. \ref{figfruit1} and Fig. \ref{figfruit2}. Fig. \ref{figfruit1} (a), (b), (c) and (d) all exhibit prominent building facades, losing the characteristics of true orthographic photos. In Fig. \ref{figfruit1}(a,4), a blurred thick edge appears on the left facade, indicating significant distortion. Similar degradation can be also found in Fig. \ref{figfruit2} (a),(b), (c). However, the Tortho-Gaussian method can successfully retain continuous linear building edges while completely eliminating the projections of building facades.

\subsubsection{Slender structures}
\begin{figure}[htbp]
\centering
\includegraphics[width=0.485\textwidth]{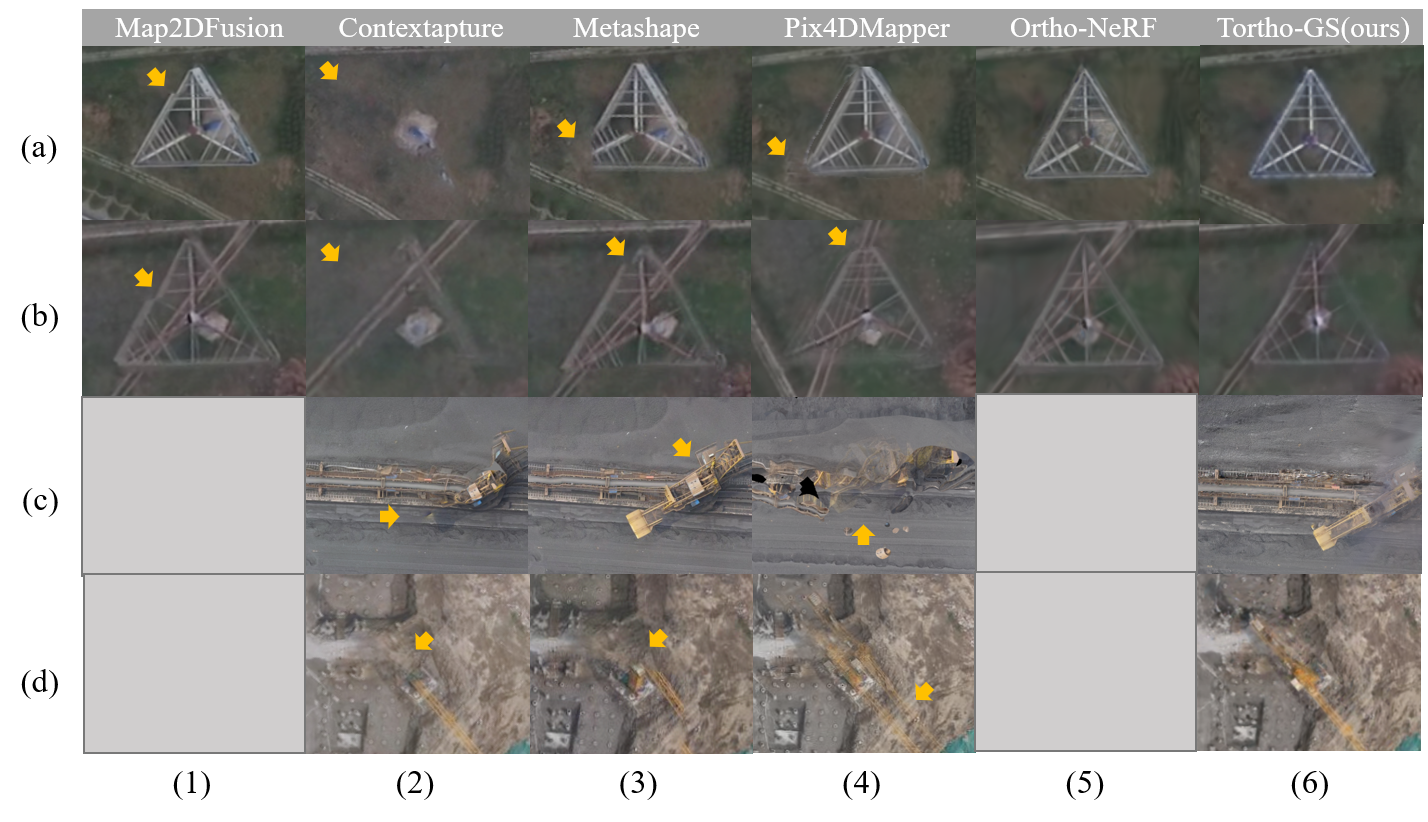}
\caption{Visual comparison of TDOMs using different methods on slender scenes. The gray block
denotes that the result is not available due to the practical re-implementation issue}
\label{figfruit3}
\end{figure}

Reconstructing slender and thin structures, such as non-rigid tree canopies, trunks, and cables, is particularly challenging due to the inherent artifacts in novel view synthesis \cite{69yu2024mip}. To address these issues, we employ the fully anisotropic Gaussian kernel functions as a super-resampling technique specifically designed to handle structures prone to aliasing \cite{70barron2023zip,71hu2023tri,72wang2023rip}. As the results of slender structures shown in Fig. \ref{figfruit3} (a), (b), (c), (d), our method outperforms other approaches in both effectively and clearly restoring the geometric details of triangular power, towers, excavators and cranes. Our TOrtho-Gaussian approach demonstrates exceptional efficacy in reconstructing slender structures and thin shells, preserving the visual integrity of true orthographic images without distortion, breakage, or noise. This highlights its capability to handle challenging scenarios involving complex, fine-scale geometries with high fidelity.
\begin{table*}[htbp]
\centering
\caption{Training Time and VRAM Usage on Phantom Dataset. Best results are in bold.}
\label{table:consumption}
\begin{tabular}{lcc|cc|cc}
\toprule
\multirow[t]{2}{*}{Dataset} & \multicolumn{2}{c}{Phantom-ieu} & \multicolumn{2}{c}{Phantom-factory} & \multicolumn{2}{c}{Phantom-npu} \\
\cmidrule(lr){2-3} \cmidrule(lr){4-5} \cmidrule(lr){6-7}
Method & Training & VRAM & Training & VRAM & Training & VRAM \\
\midrule
Vanilla 3DGS & Failed& 21.6 GB& Failed & 22.8G & Failed & 22.2G \\
TOrtho-Gaussian & \textbf{39m23s}& \textbf{10.9G} & \textbf{42m5s} & \textbf{10.2G} & \textbf{43m7s} & \textbf{11.2G} \\
\bottomrule
\end{tabular}
\end{table*}
\begin{figure}[htbp]
\centering
\includegraphics[width=0.485\textwidth]{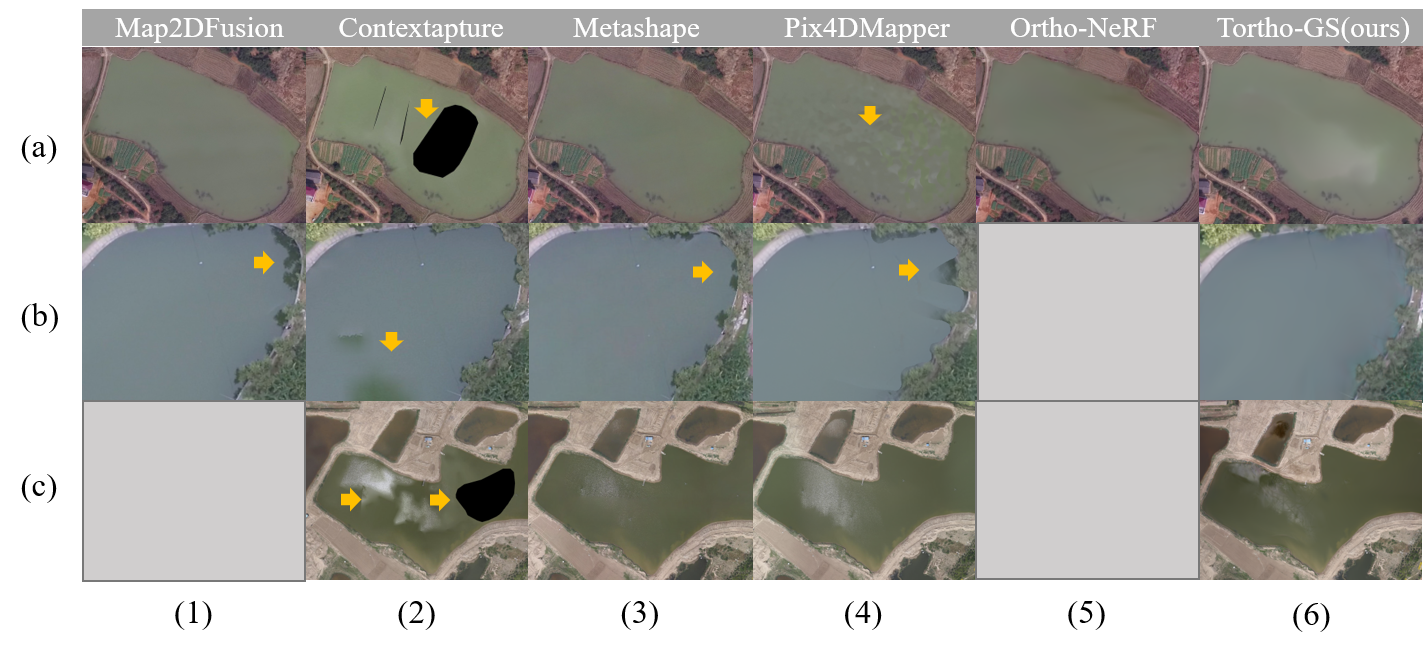}
\caption{Visual comparison of TDOMs using different methods on weak texture scenes. The gray block denotes that the result is not available due to the practical re-implementation issue}
\label{figfruit4}
\end{figure}

\subsubsection{Weak Texture}
Weak texture often occurs in generating TDOM, such as water bodies and lake surfaces, which is difficult for traditional methods to fully reconstruct, as evidenced by the ghosting in Fig. \ref{figfruit4} (a,4), (b,2) and Fig. \ref{figfruit4} (c,1), the holes in  Fig. \ref{figfruit4} (a,2) and Fig. \ref{figfruit4} (c,2), the tree reflections along the edges in Fig. \ref{figfruit4} (b,1), (b,3), and (b,4). These are all results of failed weak-texture reconstructions, due to the limitations of conventional TDOM methods. Our Gaussian splatting technique offers an effective alternative to these challenges. By accurately fitting the Gaussian field to weak-texture regions, our method ensures a continuous, smooth surface, minimizing the case of holes. The differentiable nature of the Gaussian field further enhances this continuity, while true orthographic splatting effectively eliminates artifacts like tree reflections. Additionally, the continuous Gaussian field acts as a filter, reducing erroneous texture noise and improving overall reconstruction quality.
\subsection{Quantitative Evaluation}
Due to the absence of 3D control points, two popular commercial software, Metashape and Pix4DMapper, are applied as ground truth to evaluate the relative precision of our TDOM. We measure the lengths of line segments at building corner points on the TDOM from Tortho-Gaussian, Metashape, and Pix4DMapper. Each approach's TDOM has eight groups of two line segments, and the ratio of the two line lengths is taken as an indicator for TDOM quality, which ideally should remain identical across all DTOMs. Thus, we calculate the absolute and relative errors between these ratios as Tab. \ref{table:error_comparison} lists. The average relative error and average absolute error of the ratios obtained from Tortho-Gaussian and Metashape are 0.155\% and 0.001622, respectively. Similarly, the average relative error and average absolute error between Tortho-Gaussian and Pix4DMapper are 0.126\% and 0.001772, respectively. These findings indicate that the TDOM produced by our method achieves a level of accuracy comparable to state-of-the-art commercial software, demonstrating its reliability for precise geospatial reconstruction. 

\begin{figure}[htbp]
\centering
\includegraphics[width=0.46\textwidth]{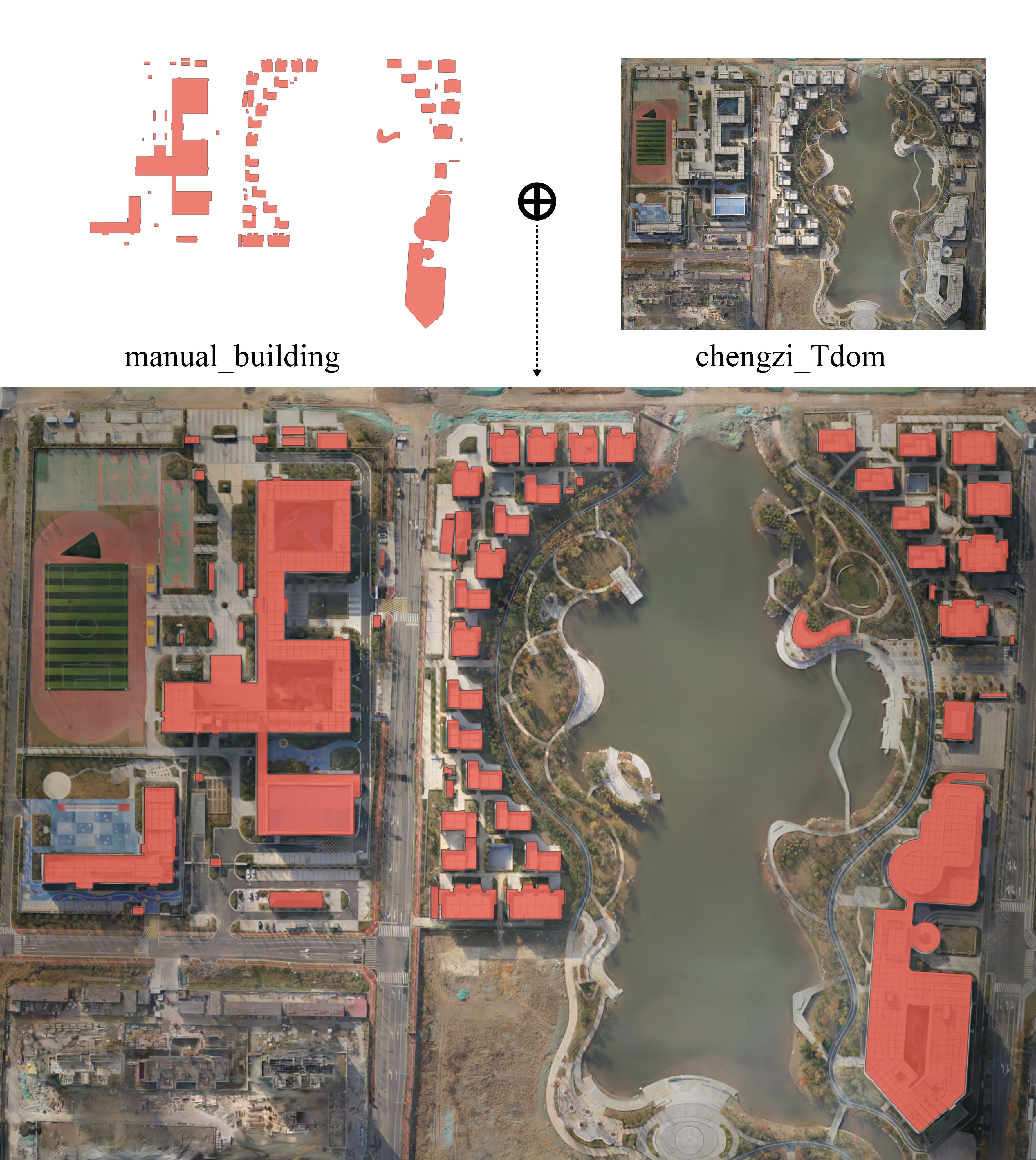}
\caption{The overlaid result with CAD map. Our generated TDOM is overlaid with an referenced CAD map produced via manual editing, the overlapping areas are highlighted with red background.}
\label{figshape}
\end{figure}

To further investigate the mapping accuracy, we conduct an overlay analysis using the Chengzi dataset by superimposing the TDOM generated by our method onto the corresponding CAD vector maps. As illustrated in Fig. \ref{figshape}, the overlay results demonstrate that our TDOM aligns precisely with the CAD maps, accurately replicating the shapes and boundaries of various features. This highlights the reliability and robustness of the Tortho-Gaussian approach in preserving geometric fidelity and ensuring accurate representation of spatial content.
\subsection{Efficiency}
We explored the optimization time and video memory cost for orthographic photos using the Tortho-Gaussian and vanilla 3DGS. Based on the Phantom3 dataset, both methods were tested with 30,000 iterations to ensure full convergence of the 3D Gaussian fields. 

Tab.\ref{table:consumption} provide the results from nine Gaussian field strips. Our proposed method can significantly reduces the cost time for reconstructing large-scale scenes. Notably, Tortho-Gaussian method consistently outperforms the vanilla 3DGS in terms of optimization time and memory efficiency when applied to large-scale scenes, demonstrating its practical value for commercial orthophoto production.

\subsection{Ablation Studies}
In this section, we conduct in-situ ablation studies to look into various aspects of Tortho-Gaussian.
\subsubsection{Spatial Resolution}
\begin{figure}[htbp]
\centering
\includegraphics[width=0.45\textwidth]{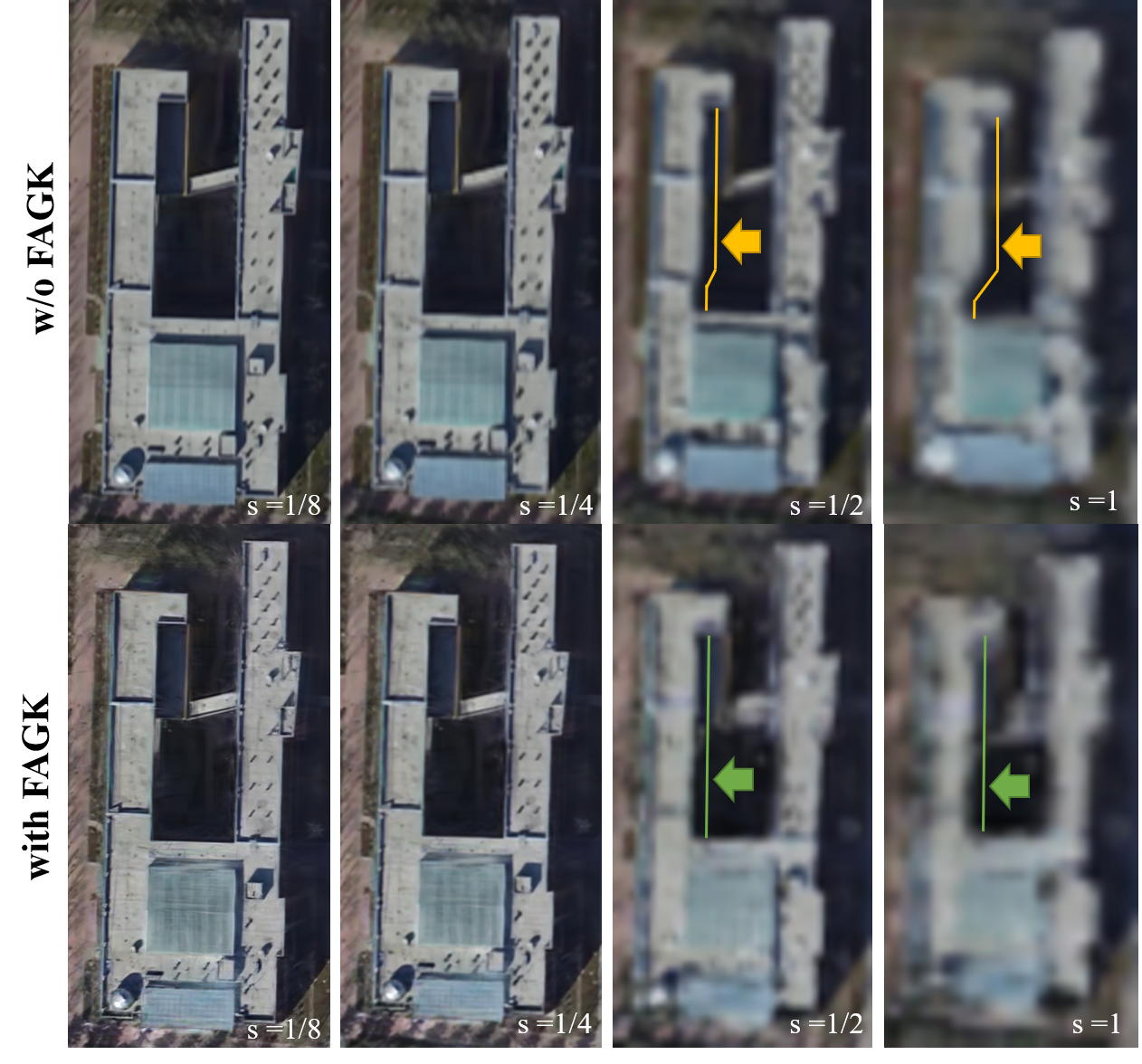}
\caption{Results of TDOM using different Spatial Resolutions(SR). From left to right, the spatial resolution decreases. As the screen space shrinks, the original 3DGS exhibits aliasing during projection onto 2D Gaussians, resulting in expansion artifacts that make the building structures appear thicker.}
\label{figaliase}
\end{figure}

As Section \ref{Ortho Splatting} explained that different spatial resolutions vary the resolution of TDOM and the number of 3D Gaussians for rendering a specific pixel. Thus, we test several set of $s_{x}$ and $s_{y}$ values to evaluate the influence of our orthogonal splatting at different spatial resolutions. Fig.  \ref{figaliase} depicts that, in general, relevant TDOMs can be accurately rendered across different spatial resolutions, indicating that our Tortho-Gaussian is capable of generating TDOM products at multiple resolutions. Nevertheless, the fidelity and details of TDOM increase as the spatial resolution becomes higher and tend to be stable, this can be explained by the fact that, for large spatial resolution, more splat 2D Gaussians are input for $\alpha$-blending which may include some noisy and adjacent yet non-relevant Gaussians. As for the higher spatial resolution, more compact 2D Gaussians are considered and the fidelity of rendered TDOM tends to stable when the spatial resolution is approaching to the real spatial resolution of the input images.

Notably, the absence of FAGK in rendering lower spatial-resolution TDOM products leads to edge dilation artifacts, particularly along the boundaries of slender structures. This issue arises from the dilation and erosion effects of Gaussian ellipsoids during rendering, which introduce high-frequency Gaussian-shaped artifacts or pronounced swelling effects when the sampling rate (e.g., focal length or camera distance) is adjusted \cite{69yu2024mip}. These artifacts can be mitigated through techniques such as multi-sampling, area sampling, or super-sampling. As illustrated in \ref{figaliase}, the incorporation of FAGK effectively address this issue by introducing refined transparency transitions to the Gaussian kernels, thereby improving the sampling precision  and significantly reducing edge-related artifacts.

\subsubsection{Regional Partitioning}
Various partitions may result in different performance on the rendering results and cost time. Tab. \ref{table:ablation_partition} compares different number of partitions, it can be found that increasing the number of partitioned regions generally results in a decline in SSIM, PSNR, and LPIPS values, indicating a degradation in image quality. This specific trend might due to that excessive partitioning can lead in additional boundary effects among various blocks. However, when the number of regions is set to one, there is an observable decrease in these values, which is most likely attributed to instable optimization and insufficient iterations for a whole large scene. As for the cost time, with four GPUs, 2$\times$2 partition is the most time efficient, because this parallel optimization strategy leverages the full computational powers of multiple GPUs and each GPU can independently optimize a block in parallel, resulting in a substantial reduction in training time. The other partition solutions (with more than 4 partitions) need to wait free available computation resources (note that, as \cite{61lin2024vastgaussian}, the cost time can be further reduced if more multiple GPUs are employed to do parallel training for multiple blocks). These findings highlight the importance of balancing the level of partitioning to optimize image quality while minimizing the introduction of artifacts.

\begin{table}[htbp]
\centering
\caption{Ablation on Data Partition Strategy.}
\label{table:ablation_partition}
\begin{threeparttable} 
\begin{tabular}{lccccc}
\toprule
\multirow{2}{*}{Scene} & \multirow{2}{*}{Partition} & \multicolumn{4}{c}{Metrics} \\
\cmidrule(lr){3-6}
& & SSIM & PSNR & LPIPS & Time \\
\midrule
\multirow{5}{*}{Phantom3-ieu} & 1 & 0.831 & 26.25 & 0.173 & 0h41m \tnote{*} \\
& 2×2 & \textbf{0.921} & \textbf{31.01} & 0.057 & \textbf{0h39m} \\
& 3×3 & 0.919 & 30.77 & \textbf{0.053} & 1h35m \\
& 4×4 & 0.902 & 28.90 & 0.061 & 2h2m \\
& 5×5 & 0.907 & 28.95 & 0.059 & 3h19m \\
\bottomrule
\end{tabular}
\begin{tablenotes}
\item[*] For this part, an modified 3DGS with optimized memory usage is applied. As section \ref{settings} explains, 30000 iterations is run for the whole scene in total. And for 2$\times$2 partitions, each subregion is also optimized by 30000 iterations.  
\end{tablenotes}
\end{threeparttable}
\end{table}

\subsubsection{Fully Anisotropic Gaussian Kernel}
Tab. \ref{table:ablation_fagk} provides the results of with/without the proposed FAGK using various degree setting (see section \ref{FAGK}). We can see that incorporating our fully anisotropic Gaussian kernels can typically improve the SSIM, PSNR, and LPIPS values, and higher degree can further enhance the rendering results. This improvement is attributed to the FAGK’s ability to enable super-sampling, which enhances reflectivity sensitivity and provides better anti-aliasing effects.
\begin{table}[htbp]
\centering
\caption{Ablation Study on Fully Anisotropic Gaussian Kernels.}
\label{table:ablation_fagk}
\begin{tabular}{lcccc}
\toprule
Degree setting & Model setting & SSIM & PSNR & LPIPS \\
\midrule
\multirow{2}{*}{1} & w/o FAGK & 0.813 & 18.22 & 0.285 \\
                    & Full, (Ours) & 0.806 & 19.27 & 0.296 \\
\midrule
\multirow{2}{*}{2} & w/o FAGK & 0.867 & 23.18 & 0.223 \\
                    & Full, (Ours) & 0.867 & 23.80 & 0.224 \\
\midrule
\multirow{2}{*}{3 (full)} & w/o FAGK & 0.868 & 23.33 & \textbf{0.218} \\
                           & Full, (Ours) & \textbf{0.870} & \textbf{25.10} & 0.224 \\
\bottomrule
\end{tabular}
\end{table}

\section{Conclusion}
In this paper, we presented a novel approach for generating True Digital Orthophoto Maps (TDOMs) by utilizing the 3D Gaussian Splatting (3DGS) technique, namely, Tortho-Gaussian. Our Tortho-Gaussian can bypass the traditional TDOM generation difficulties of occlusion detection and distortion correction, enabling a direct and efficient production of high-quality TDOM. Our throughout experimental results demonstrate that, Tortho-Gaussian not only outperforms existing commercial software in terms of TDOM precision and quality, but also provides a scalable solution for large-scale scene reconstruction. By dividing scenes into manageable segments and optimizing them independently, our method addresses the computational challenges associated with large datasets and achieves efficient memory usage without compromising rendering quality.

In the future, two potential relevant directions could be explored: first, to handle even more large-scale scenes (e.g., city-level), it is necessary to light the 3DGS and adopt a more reasonable partition strategy to minimize the thread wait among blocks. Second, we would like to incorporate other semantic information (e.g., segment anything model \cite{kirillov2023segment}) and depth (depth anything \cite{yang2024depth}) into our Tortho-Gaussain to further improve the quality of TDOM. 

\ifCLASSOPTIONcaptionsoff
  \newpage
\fi

\bibliographystyle{IEEEtran}
\bibliography{references}
\begin{IEEEbiography}[{\includegraphics[width=1.0in,height=1.25in,clip,keepaspectratio]{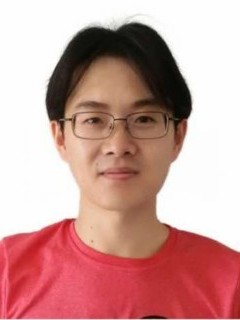}}] {Dr.-Ing Xin Wang}
(IEEE Member) received his bachelor and master in surveying and mapping from school of Geodesy and Geomatics, Wuhan University, China, in 2013 and 2016, respectively. In 2021, he obtained Doctor of Engineering in photogrammetry and remote sensing from Leibniz university Hannover, Germany. He is currently an assistant professor in Wuhan University, whose research interests are computer vision, deep learning in applied photogrammetry etc.
\end{IEEEbiography}
\begin{IEEEbiography}[{\includegraphics[width=1.0in,height=1.25in,clip,keepaspectratio]{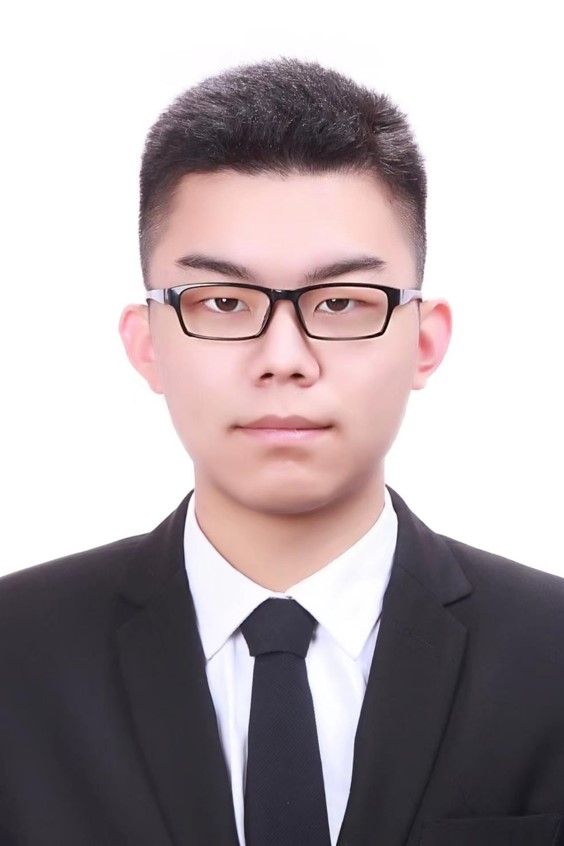}}] {Wendi Zhang}
received his bachelor's degree in Geomatics Engineering from China University of Petroleum in 2023 and is currently pursuing a master's degree at Wuhan University, Wuhan, China. His research interests include the application of machine learning and computer vision in photogrammetry and implicit reconstruction.
\end{IEEEbiography}
\begin{IEEEbiography}[{\includegraphics[width=1in,height=1.25in,clip,keepaspectratio]{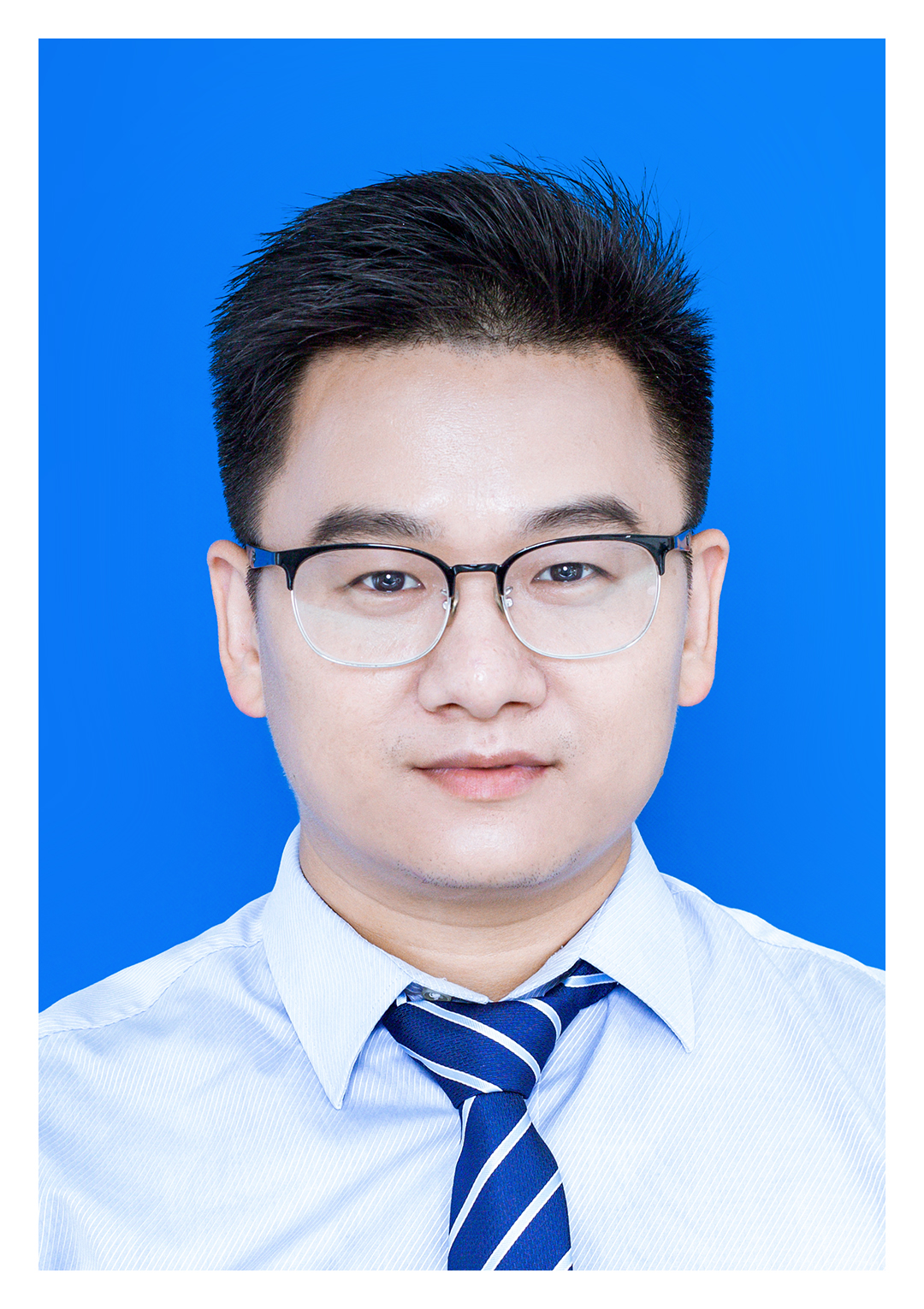}}] {Associated Prof. Hong Xie}
received the B.S., M.S., and Ph.D. degrees in photogrammetry and remote sensing from Wuhan University, Wuhan, China, in 2007, 2009, and 2013, respectively. 
He is currently an Associate Professor with the School of Geodesy and Geomatics, Wuhan University. His research interests include target detection based on image deep learning, point cloud data quality improvement, point cloud information extractionand model reconstruction, mobile mapping, and surveying.
\end{IEEEbiography}
\begin{IEEEbiography}[{\includegraphics[width=1in,height=1.25in,clip,keepaspectratio]{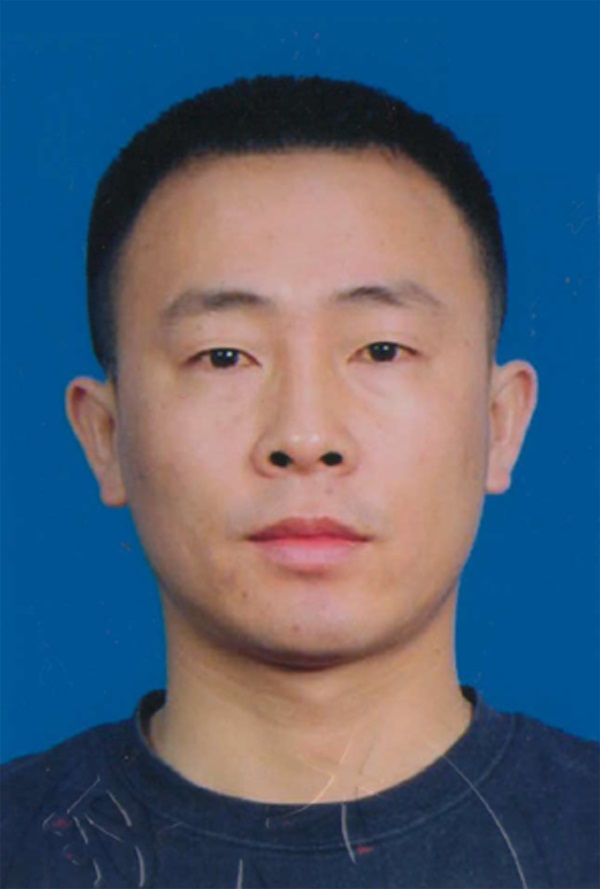}}] {Dr. Haibin Ai}
received his Ph.D. in Photogrammetry and Remote Sensing from Wuhan University in 2009. He is currently a Researcher at the Chinese Academy of Surveying and Mapping, serving as the Head of the Aerial and Space-borne Photogrammetry Research Group. He leads a team dedicated to research in digital photogrammetry for aerial and space borne platforms, remote sensing image processing, computer vision, and 3D reconstruction.
\end{IEEEbiography}
\begin{IEEEbiography}[{\includegraphics[width=1in,height=1.8in,clip,keepaspectratio]{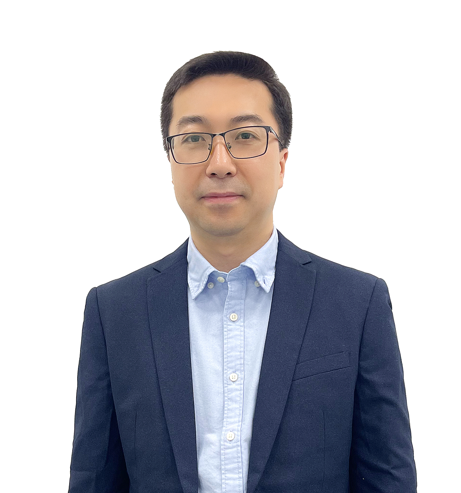}}] {Prof. Qiangqiang Yuan}
(IEEE Member) received the B.S. degree in surveying and mapping engineering and the Ph.D. degree in photogrammetry and remote sensing from Wuhan University, Wuhan, China, in 2006 and 2012, respectively. In 2012, he joined the School of Geodesy and Geomatics, Wuhan University, where he is a Professor. He has published more than 100 research articles, including more than 80 peer-reviewed articles in international journals, such as Nature Communications, Remote Sensing of Environment, ISPRS Journal of Photogrammetry and Remote Sensing, IEEE TRANSACTIONS ON IMAGE PROCESSING, and IEEE TRANSACTIONS ON GEOSCIENCE AND REMOTE SENSING. His research interests include image reconstruction, remote sensing image processing and application, and data fusion. Dr. Yuan was a recipient of the Youth Talent Support Program of China in 2019, the Top-Ten Academic Star of Wuhan University in 2011, and the recognition of Best Reviewer of IEEE GEOSCIENCE AND REMOTE SENSING LETTERS in 2019. In 2014, he received the Hong Kong Scholar Award from the Society of Hong Kong Scholars and China National Postdoctoral Council. He is an associate editor of five international journals and has frequently served as a referee for more than 40 international journals for remote sensing and image processing.
\end{IEEEbiography}
\begin{IEEEbiography}[{\includegraphics[width=1in,height=1.25in,clip,keepaspectratio]{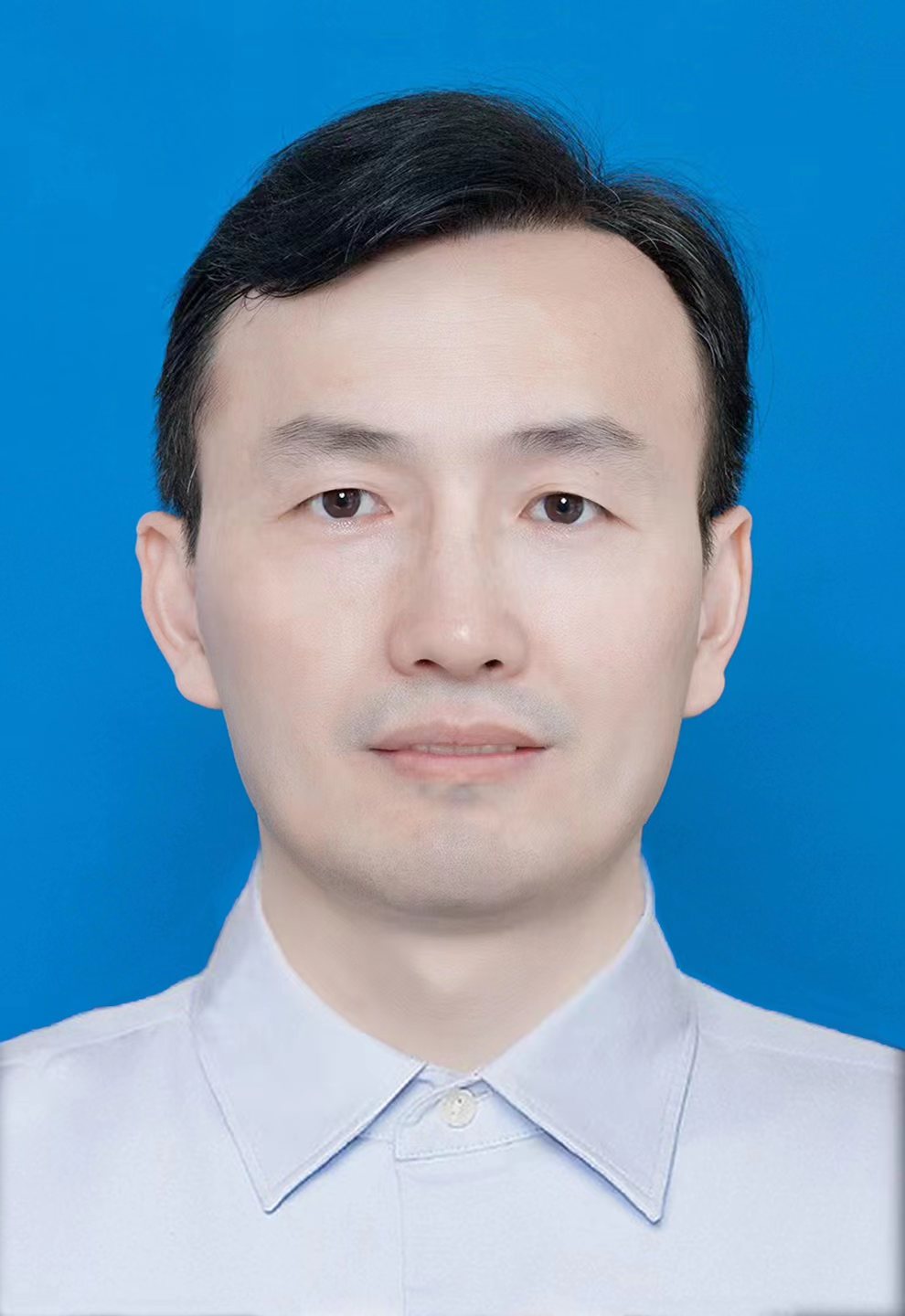}}] {Prof. Zongqian Zhan}
(IEEE Member) received the M.A.Eng. and Ph.D. degrees in photogrammetry and remote sensing from Wuhan University, Wuhan, China, in 2003 and 2007, respectively. He is currently a full Professor with the School of Geodesy and Geomatics, Wuhan University. His research interests include camera calibration, close-range and UAV photogrammetry, oblique photogrammetry, deep learning, and remote sensing.
\end{IEEEbiography}
\end{document}